\documentclass{article}

\usepackage{microtype}
\usepackage{graphicx}
\usepackage{subfigure}
\usepackage{booktabs} 

\usepackage{hyperref}

\usepackage[accepted]{icml2025}

\usepackage{amsmath}
\usepackage{amssymb}
\usepackage{mathtools}
\usepackage{amsthm}

\usepackage[capitalize,noabbrev]{cleveref}

\theoremstyle{plain}

\theoremstyle{definition}

\theoremstyle{remark}

\usepackage[textsize=tiny]{todonotes}

\icmltitlerunning{AL-Guided Seq2Seq VAE for Multi-target Generation}

\begin{document}

\twocolumn[
\icmltitle{Active Learning-Guided Seq2Seq Variational Autoencoder\\
            for Multi-target Inhibitor Generation}




\begin{icmlauthorlist}
\icmlauthor{Júlia Vilalta-Mor}{yyy,hhh}
\icmlauthor{Alexis Molina}{comp}
\icmlauthor{Laura Ortega Varga}{xxx,zzz}
\icmlauthor{Isaac Filella-Merce}{yyy}\textsuperscript{*}
\icmlauthor{Victor Guallar}{yyy,comp,sch}\textsuperscript{*}
\end{icmlauthorlist}

\icmlaffiliation{yyy}{Barcelona Supercomputing Center (BSC), Plaça d’Eusebi Güell, 1-3, 08034, Barcelona, Spain}
\icmlaffiliation{hhh}{PhD program in Biotechnology, Faculty of Pharmacy and Food Sciences,
University of Barcelona (UB), 08028 Barcelona, Spain}
\icmlaffiliation{comp}{Nostrum Biodiscovery S.L., Av. de Josep Tarradellas, 8-10, 3-2, 08029, Barcelona, Spain}
\icmlaffiliation{xxx}{Alzheimer's Research UK Oxford Drug Discovery Institute, Centre for Medicines Discovery, Nuffield Department of Medicine, Oxford, UK}
\icmlaffiliation{zzz}{Centre for Medicines Discovery, Nuffield Department of Medicine Research Building (NDMRB), University of Oxford, Oxford, UK}
\icmlaffiliation{sch}{ICREA, Pg. Lluis Companys 23, 08010, Barcelona, Spain}

\icmlcorrespondingauthor{Victor Guallar}{victor.guallar@bsc.es}
\icmlcorrespondingauthor{Isaac Filella-Merce}{ifilell1@bsc.es}

\icmlkeywords{Generative Models, Active Learning, Seq2Seq, Variational Autoencoder, Multi-target, Drug Design, Pan-inhibitors, Coronaviruses}

\vskip 0.3in
]



\printAffiliationsAndNotice{}  

\begin{abstract}
Simultaneously optimizing molecules against multiple therapeutic targets remains a profound challenge in drug discovery, particularly due to sparse rewards and conflicting design constraints. We propose a structured active learning (AL) paradigm integrating a sequence-to-sequence (Seq2Seq) variational autoencoder (VAE) into iterative loops designed to balance chemical diversity, molecular quality, and multi-target affinity. Our method alternates between expanding chemically feasible regions of latent space and progressively constraining molecules based on increasingly stringent multi-target docking thresholds. In a proof-of-concept study targeting three related coronavirus main proteases (SARS-CoV-2, SARS-CoV, MERS-CoV), our approach efficiently generated a structurally diverse set of pan-inhibitor candidates. We demonstrate that careful timing and strategic placement of chemical filters within this active learning pipeline markedly enhance exploration of beneficial chemical space, transforming the sparse-reward, multi-objective drug design problem into an accessible computational task. Our framework thus provides a generalizable roadmap for efficiently navigating complex polypharmacological landscapes.
\end{abstract}

\section{Introduction}
\label{introduction}

Generative models (GMs), a machine learning approach widely applied in fields such as text generation, can be effectively used in drug discovery by treating molecular representations (e.g., SMILES strings) as a chemical language \cite{ahmad_generative_2024}. These models enable the design of truly novel molecules by navigating unexplored regions of chemical space that are inaccessible to traditional screening methods reliant on predefined molecular libraries \cite{tang_survey_2024}. This is particularly evident in ultra-large chemical libraries \cite{tingle_zinc-22free_2023, noauthor_real_nodate}, which, despite containing tens of billions of compounds, often lack diversity due to their construction from repeated combinations of limited sets of building blocks and chemical reactions \cite{neumann_benchmark_2025}. In contrast, GMs learn underlying chemical patterns from data and generate new molecular structures beyond the scope of existing libraries \cite{zhavoronkov_deep_2019, grisoni_combining_2021, korshunova_generative_2022, swanson_generative_2024}.

Among GMs, Variational Autoencoders (VAEs) have shown particular promise. Their minimal backbone consists of an encoder, which compresses data into a lower-dimensional latent space, and a decoder, which reconstructs the original data from this representation. Sequence-to-Sequence (Seq2Seq) VAEs are especially effective in learning latent representations of molecules, enabling a controlled sampling of new molecules with optimised physicochemical and pharmacological properties \cite{gomez-bombarelli_automatic_2018, filella-merce_optimizing_2023}. Building on this directed generation, the integration of methods such as reinforcement learning (RL) \cite{sheikholeslami_druggen_2025} and active learning (AL) into GMs workflows has further improved their ability to guide molecular generation toward specific objectives. For instance, by iteratively guiding the model with feedback from molecular modelling simulations such as docking, these workflows can prioritise molecules. Specifically, those with enhanced affinity towards a given target \cite{filella-merce_optimizing_2023}. 

GM workflows have primarily focused on generating molecules with affinity for single targets. In recent years, increasing attention has been given to multi-target drug discovery, \emph{i.e.} molecules with simultaneous affinity to multiple targets. Molecular generation aimed at multi-target inhibition could pave the way for the development of polypharmacological drugs, offering a powerful new approach to treating complex diseases such as cancer \cite{cichonska_ai_2024, isigkeit_automated_2024}. Similarly, multi-target generation also aligns with the concept of pan-inhibitors, where a single molecule exerts therapeutic effects across different organisms by targeting homologous proteins \cite{shahhamzehei_silico_2022, huang_new_2023}. \citet{liu_drugex_2021} and \citet{munson_novo_2024} explored the use of RL to construct a multi-target GM. Their goal was to steer the generation toward predefined desirable properties, by rewarding generated molecules that successfully meet the specified criteria. However, these approaches, like many RL-based methods \cite{olivecrona_molecular_2017, blaschke_memory-assisted_2020, sheikholeslami_druggen_2025, haddad_targeted_2025}, face challenges due to the sparse reward problem. This is particularly relevant in the case of target affinity, where the large imbalance between inactive and active molecules hinders the model's ability to learn effective strategies for obtaining the desired reward \cite{korshunova_generative_2022}. As an alternative, AL can be used to iteratively select only active molecules to retrain the GM. In addition, AL can help overcome challenges commonly faced by standalone GMs, including poor target engagement under low-data regimes \cite{van_tilborg_traversing_2024}, limited synthesizability of the generated molecules, and the generalization outside the training set \cite{gangwal_current_2024,loeffler_optimal_2024,kyro_chemspaceal_2024}.

In this paper, we propose a multi-target generative workflow aimed at generating molecules with predicted simultaneous affinity to multiple targets. As a test case, we evaluated our multi-target generative workflow on the design of a pan-inhibitor targeting the main protease (Mpro) of diverse coronaviruses: SARS-CoV-2, SARS-CoV, and MERS-CoV \cite{shahhamzehei_silico_2022, huang_new_2023}.

\section{Methods}
\label{methods}

\subsection{Multi-target Generative Workflow}

The multi-target generative workflow builds upon the concept of single-target molecular generation. This involves generating molecules with predicted affinity for a single target, and extends its application to multiple biological targets.

The workflow starts by training the generator, a Seq2Seq VAE \cite{gomez-bombarelli_automatic_2018, gupta_generative_2018},  with a general dataset of molecules represented in text-based format (SMILES). This first training teaches the VAE the underlying grammar, thereby enabling the generation of chemically feasible molecules. Subsequently, the Seq2Seq VAE is fine-tuned with a fixed specific dataset of molecules with known affinity towards multiple targets, without requiring affinity to all targets simultaneously (\cref{appendixA}). By doing so, the generation is biased towards molecules with affinity to the multiple targets.

\begin{figure*}[ht]
\vskip 0.2in
\begin{center}
\centerline{\includegraphics[width=1.7\columnwidth]{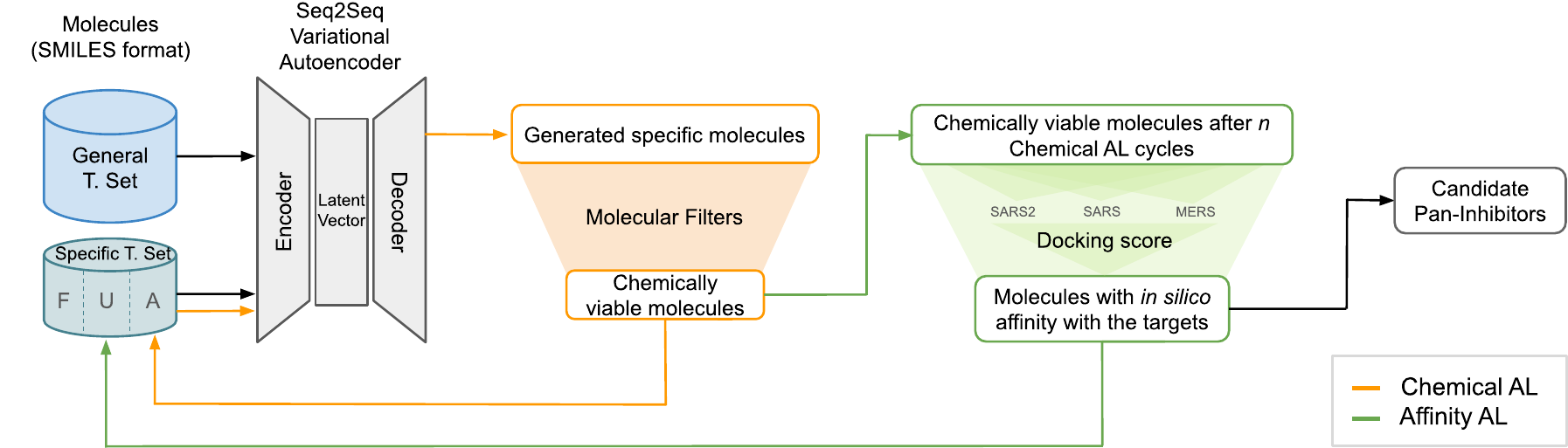}}
\caption{The multi-target generative workflow. The two-level AL workflow is illustrated with arrows of different colours: the first-level cycle, Chemical AL, is shown in orange, while the second-level cycle, Affinity AL, is shown in green. The specific training set stands for F: fixed, U: updated, and A: accumulated.}
\label{figure1}
\end{center}
\vskip -0.2in
\end{figure*}

Once the VAE is pretrained, the multi-target generative workflow initiates an iterative process of molecular generation and refinement through a two-level AL workflow. The first level AL cycle, or Chemical AL, promotes molecules based on physicochemical properties. The second level AL cycle, or the Affinity AL, promotes molecules based on simultaneous predicted affinity to multiple targets (hereafter referred to as multi-target affinity) (\cref{figure1}).

The two-level AL workflow begins by performing the Chemical AL cycle $n$ times. In each cycle, new molecules are generated and filtered based on the presence of undesired structural motifs and chemoinformatic predictor thresholds. The resulting molecules are then used to fine-tune the VAE, starting from the general training weights. This fine-tuning utilises all accumulated molecules from previous and current Chemical AL cycles (the accumulated specific dataset), as well as molecules in the fixed specific dataset. After completing the $n$ Chemical AL cycles, an Affinity AL cycle is conducted, in which all molecules in the accumulated specific dataset are filtered based on their multi-target affinity. The VAE is then fine-tuned, starting from the general training weights, using the accumulated molecules from previous and current Affinity AL cycles (the updated specific dataset) along with those from the fixed specific dataset. After $m$ Affinity AL cycles, this two-level AL approach enables the progressive refinement of molecule generation, first by enforcing favourable chemical properties, and then by optimising for multi-target affinity (\cref{figure1}). Currently, the workflow runs on a single GPU, and each Affinity AL cycle can take approximately 18 hours to complete (\cref{appendixG}).

\subsection{Seq2Seq Variational Autoencoder}

We implemented a Seq2Seq VAE architecture that processes molecules as SMILES sequences, one-hot encoded with a vocabulary size of $D=50$. The encoder consists of a single LSTM layer that processes the input sequence $x=(x_1, \text{…} , x_T)$ producing a final hidden state $h_{T}$, which is then passed through a fully connected layer with 256 units and ReLU activation:

\begin{align*}
    h'=ReLU(W^{enc}h_T+b^{enc}) \tag{1}
\end{align*}

From $h’$, the VAE models the latent space as a probabilistic Gaussian distribution with mean vector $\mu \in \mathbb{R}^{128}$ and log-variance vector $log \text{ } \sigma^2 \in \mathbb{R}^{128}$ computed as:

\begin{align*}
    \mu = W_{\mu}h' + b_{\mu} \text{, } log\sigma^2=W_{\sigma}h'+b_{\sigma} \tag{2}
\end{align*}

where $W_{\mu}, W_{\sigma} \in \mathbb{R}^{128 \times256}$ and $b_{\mu}, b_{\sigma} \in \mathbb{R}^{128}$.
The latent vector $z$ is sampled using the reparametrization trick:
\begin{align*}
    z=\mu + \sigma \odot \epsilon, \epsilon \in \mathcal{N}(0,I), \sigma=e^{\frac{1}{2}log\sigma^2 \tag{3}}
\end{align*}

where $\epsilon$ is a random noise vector sampled from a standard normal distribution. 

The decoder reconstructs the sequence by transforming $z$ through a fully connected layer with 256 units and ReLU activation, which initializes the hidden state of a decoder LSTM. Finally, the decoder logits are passed to a softmax layer to obtain the output sequence.

During training, the VAE learns to reconstruct the input sequence $x$ from a compressed latent representation $z$, while also shaping the latent space so that similar inputs map to nearby latent vectors. This is achieved by maximizing the Evidence Lower Bound (ELBO), a composite loss function that combines the Reconstruction loss and the Kullback-Leibler (KL) divergence \cite{odaibo_tutorial_2019}.

The KL‐divergence term in the VAE’s loss regularizes the approximate posterior $q(z\mid x)$ toward the standard normal prior $p(z)=\mathcal{N}(0,I)$, yielding a continuous, densely populated latent space of valid molecular embeddings. To generate new molecules, one samples $z\sim N(0,I)$. 
The sampled vector $z$ is then passed through a linear (fully connected) layer to initialize the decoder’s LSTM hidden (and, if applicable, cell) state. From there, the decoder operates autoregressively: at each time step, it takes the embedding of the previously generated token (and optionally $z$ again), updates its hidden state, and outputs a probability distribution over the next token. This process repeats until the end‐of‐sequence token is emitted or a predefined maximum length is reached, yielding a complete SMILES string.

\subsection{Chemical Active Learning Cycle}

The goal of the Chemical AL cycle is to guide molecular generation towards synthesisable, drug-like molecules that meet user-defined requirements for structural variability. A key concern in molecular generative AI is the limited synthetic accessibility and chemical quality of the generated molecules. To address this, the Chemical AL cycle starts by applying a substructure filter to remove generated molecules containing undesirable structural motifs that could hinder their synthesis and progression in a drug design campaign. Retention of such motifs could lead to their accumulation due to the iterative nature of the workflow (error propagation). Then, the filtered molecules are evaluated against three cheminformatic predictors designed to improve drug-likeness, further enhance synthetic accessibility, and guide molecular generation based on structural variability. These filtering predictors are: (1) Quantitative Estimate of Drug-likeness (QED) \cite{bickerton_quantifying_2012}, ranging from 0 (least drug-like) to 1 (most drug-like), (2) Synthetic Accessibility (SA) score \cite{ertl_estimation_2009}, ranging from 1 (easy to synthesise) to 10 (difficult to synthesise), and (3) Tanimoto similarity (TA) \cite{bajusz_why_2015} against the specific dataset, computed using Morgan4 fingerprints, with values from 0 (no similarity) to 1 (identical molecules). Finally, medicinal chemistry filters are applied using 4 SMARTS-based catalogues to filter generated molecules containing undesirable structural motifs associated with toxicity, promiscuity, or poor pharmacokinetics. These catalogues (found in the Python package RDKit \cite{landrum_rdkitrdkit_2025}) include PAINS \cite{baell_new_2010}, Brenk \cite{brenk_lessons_2008}, NIH \cite{doveston_unified_2014,jadhav_quantitative_2010}, and CHEMBL \cite{walters_patwaltersrd_filters_2025}. Molecules that passed all filters are added to the accumulated specific dataset.

\subsection{Affinity Active Learning Cycle}

Affinity AL cycles are applied to guide molecular generation toward molecules with multi-target affinity. Ligand–protein affinities are predicted by docking molecules from the accumulated specific dataset to the multiple protein targets (\cref{appendixB,appendixC}). A key feature of this step is the use of two complementary docking score thresholds to filter molecules: (1) a global threshold computed as the mean docking score across the multiple targets, and (2) individual thresholds for each target, to prevent the global score from masking poor affinity to any single target. Molecules that satisfy all affinity thresholds are transferred to the updated specific dataset.

To progressively enhance affinity, a linear decay strategy is implemented, where docking score thresholds are decreased in each Affinity AL cycle (noting that lower docking scores correspond to higher predicted affinities). Let $T_g^{(i)}$ be the global threshold and $T_{ind}^{(i)}$ be the individual threshold at cycle $i$, with $\delta > 0$ being the decay rate. The thresholds are updated according to:

\begin{align*}
T_g^{(i+1)} &=
\begin{cases}
T_g^{(i)} - \delta, & \text{if } N^{(i)} \geq N_{\min} \tag{4}\\
T_g^{(i)}, & \text{otherwise}
\end{cases} \\[1.5ex]
T_{\text{ind}}^{(i+1)} &=
\begin{cases}
T_{\text{ind}}^{(i)} - \delta, & \text{if } N^{(i)} \geq N_{\min} \tag{5}\\
T_{\text{ind}}^{(i)}, & \text{otherwise}
\end{cases}
\end{align*}

where $N^{(i)}$ is the number of molecules passing the thresholds at cycle $i$, and $N_{min}$ is the minimum number of molecules required to continue decreasing the thresholds. This decay is conditional upon the retention of at least $N_{min}$ molecules meeting both threshold criteria; otherwise, the thresholds are maintained.

A stopping patience parameter of $p$ is introduced, stopping the generative workflow if thresholds cannot be lowered in $p$ consecutive Affinity AL cycles: 

\begin{align*}
C =
\begin{cases}
0, & \text{if } T_g^{(i+1)} < T_g^{(i)} \text{ or } T_{\text{ind}}^{(i+1)} < T_{\text{ind}}^{(i)} \tag{6}\\
C + 1, & \text{otherwise}
\end{cases}
\end{align*}

where $C$ is a counter for consecutive cycles without threshold decay, stopping the generation process when $C \ge p$.

\section{Results}

The multi-target generative workflow was evaluated on the design of a coronavirus pan-inhibitor. Specifically, the goal was to generate a molecule capable of inhibiting the main protease (Mpro) of three distinct coronaviruses: SARS-CoV-2, SARS-CoV, and MERS-CoV. Mpro is a well-established antiviral target due to its essential role in viral replication. Its high conservation across coronavirus species (see \cref{appendixB}) makes it an ideal test case for our multi-target generative workflow.

We designed a case-specific generation pipeline with two main goals: to enhance molecular diversity and to progressively improve multi-target affinity toward the three Mpro targets. The pipeline starts with an initial Affinity AL cycle of 40 Chemical AL cycles, aimed at maximising molecular diversity through a TA threshold $<$ 0.4. This is followed by shorter Affinity AL cycles, each with 10 Chemical AL cycles and a relaxed TA threshold $<$ 0.6, intended to improve multi-target affinity. These shorter Affinity AL cycles enable more frequent filtering based on multi-target affinity, with effectiveness further increased through progressively decaying docking score thresholds across successive cycles ($N_{min}=50$, $\delta=0.1 kcal/mol$, and $p=3$). All Affinity AL cycles use thresholds of QED $\ge$ 0.8 and SA $\le$ 3 to favour drug-like, synthetically accessible molecules, along with starting docking score thresholds of -7.5 kcal/mol (global) and -7.0 kcal/mol (individual).

\begin{figure*}[!ht]
\vskip 0.2in
\begin{center}
\includegraphics[width=2\columnwidth]{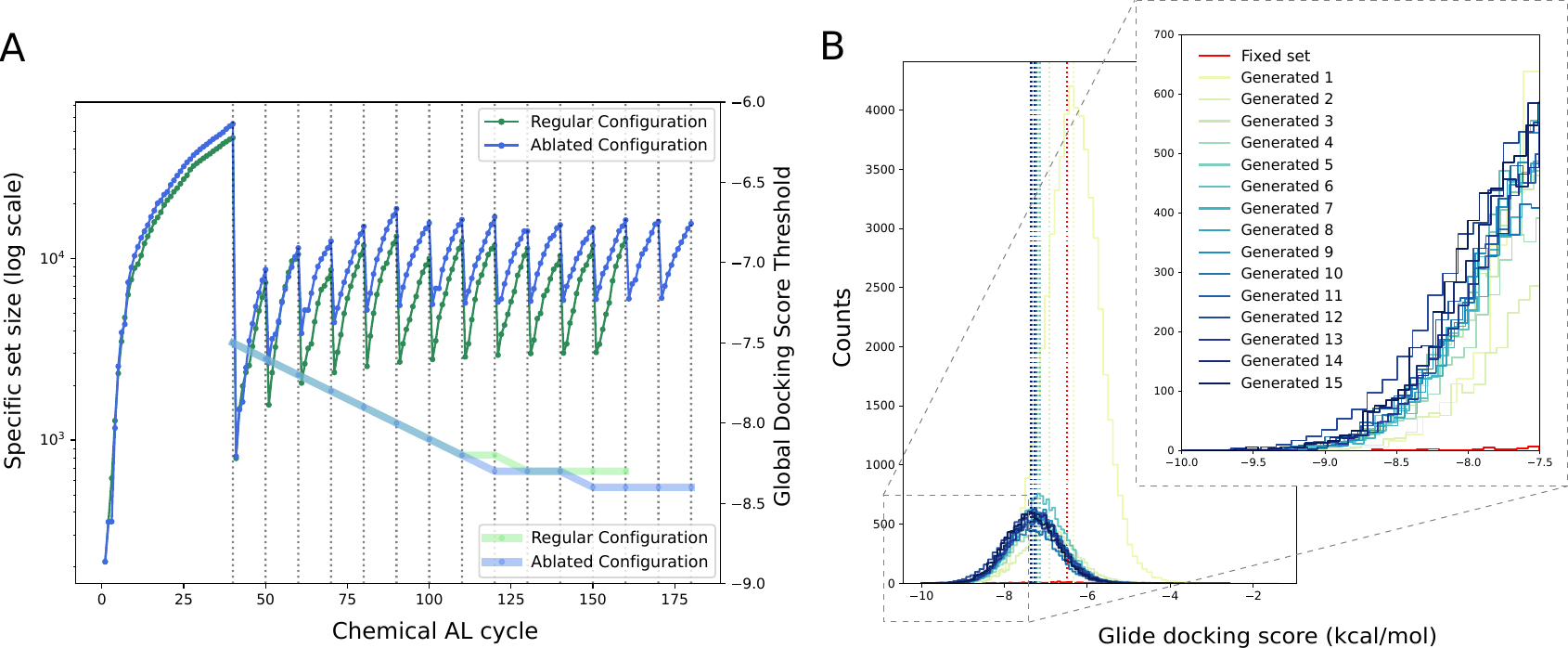}
\caption{A) Evolution of specific set size (logarithmic scale) after each Chemical AL cycle for both regular and ablated configurations. Vertical dotted lines mark the end of each Affinity AL cycle. The secondary vertical axis represents the global docking score threshold applied to each Affinity AL cycle. The initial point represents the fixed specific set size. Points between two dotted vertical lines represent the evolution of the cumulative specific set size. The drop in size at the vertical lines corresponds to the Affinity AL filtering, highlighting the drop in size of the updated specific set. B) Global docking scores histograms of generated molecules in the ablated configuration across Affinity AL cycles compared to the histogram of the fixed specific dataset. The panel with a zoom-in shows the enrichment of generated molecules in the range of high multi-target affinity.}
\label{figure2}
\end{center}
\vskip -0.2in
\end{figure*}

\subsection{Generation performance under two configurations: Regular vs Ablated}

Due to concerns that excessive filtering during the Chemical AL cycle could constrain molecular generation, we conducted two parallel generative cases. The first followed the full multi-target workflow as described in the Methods section (regular configuration). The second, serving as an ablation study, excluded SMARTS-based filtering catalogues to evaluate their impact within the Chemical AL cycles (ablated configuration).

Both generative configurations exhibited similar trends in terms of the number of generated molecules and progression through the Chemical and Affinity AL cycles (\cref{figure2}A). Due to the reduced filtering of the ablated configuration, a total of 15 AL Affinity cycles were completed before reaching the stopping parameter p, with final docking thresholds of -8.5 kcal/mol (global) and -8 kcal/mol (individual). In contrast, the regular configuration, which included the SMARTS filters within, plateaued earlier,  stopping after 13 AL Affinity cycles with slightly less stringent final thresholds of -8.3 kcal/mol (global) and -7.8 kcal/mol (individual). 

\cref{figure3} (\cref{appendixD}) compares the global docking scores distributions of the generated molecules for the regular and the ablated configuration. In the ablated configuration, the distribution shifts leftward, displaying a higher molecular count in the high-affinity region (lower range of docking score), indicating better predicted multi-target affinity of the generated molecules. Importantly, the total number of generated molecules passing all docking score thresholds and thus, accumulated in the updated specific dataset, was consistently higher in the ablated configuration, even after applying SMARTS filtering post-generation (\cref{table1}). Specifically, we observed a 1.35-fold increase in molecular count under the -7.5 kcal/mol global and -7 kcal/mol individual thresholds, which increased to a 3-fold increase when applying the more stringent -9 kcal/mol global and -8 kcal/mol individual thresholds. This suggests that deferring such filters from the AL workflow and placing them as a post-generative filter offers a more permissive yet still chemically relevant approach to molecular generation.

\subsection{Candidate pan-inhibitors}

Given the enhanced capacity of the ablated configuration, followed by post-generation SMARTS filtering, to generate molecules with multi-target affinity, we focused the remainder of our analysis on this configuration. \cref{figure2}B presents the distribution of the global docking scores for the generated molecules across the different Affinity AL cycles, as well as for those in the fixed specific dataset. Notably, the zoom-in panel reveals a consistent increase in the number of molecules in the lower docking score range with successive AL cycles. This effect is driven by the iterative nature of our AL workflow and the linear decay strategy applied to the docking score thresholds. As expected, we observe a marked enrichment of low-scoring molecules compared to the fixed set, which was not designed for multi-target affinity.

From this point onwards, we consider as a candidate pan-inhibitor any generated molecule with an individual docking score threshold of -8 kcal/mol for each target. Setting threshold values of -8 kcal/mol (global) and -8 kcal/mol (individual) yields a total of 310 molecules in the regular configuration, and a total of 650 molecules in the ablated configuration after applying SMARTS filtering (\cref{table1}). Notably, only one molecule in the fixed specific set meets these docking score thresholds, highlighting the ability of our multi-target generative workflow to generalise toward the design of multi-target inhibitors.

\begin{table*}
\caption{Percentage identity derived from the structure-based sequence alignment just on the catalytic site of the Mpro for SARS-CoV-2 (SARS2), SARS-CoV (SARS), and MERS-CoV (MERS).}
\label{table1}
\vskip 0.15in
\begin{center}
\begin{small}
\begin{sc}
\begin{tabular}{ p{2cm} p{2cm} | p{2cm} p{2.5cm} p{2.5cm} p{3cm}}
\hline
 \multicolumn{2}{c}{Docking Thresholds} &  \multicolumn{4}{c}{Number of Compounds}\\
 \hline
\toprule
\centering Global & Individual & Fixed Specific Set & Generated Regular Config. & Generated Ablated Config. & Ablated Config. After Filtering \\
\midrule
\centering-7.5 & -7 & \multicolumn{1}{c}{10} & \multicolumn{1}{c}{15,245} & \multicolumn{1}{c}{32,101} & \multicolumn{1}{c}{20,661} \\
\centering-8   & -7.5 & \multicolumn{1}{c}{2}  & \multicolumn{1}{c}{1,591}  & \multicolumn{1}{c}{4,911}  & \multicolumn{1}{c}{3,324} \\
\centering-8   & -8   & \multicolumn{1}{c}{1}  & \multicolumn{1}{c}{310\textsuperscript{*}}   & \multicolumn{1}{c}{932}    & \multicolumn{1}{c}{650\textsuperscript{*}} \\
\centering-8.5 & -8   & \multicolumn{1}{c}{1}  & \multicolumn{1}{c}{58}     & \multicolumn{1}{c}{237}    & \multicolumn{1}{c}{168} \\
\centering-9   & -8   & \multicolumn{1}{c}{0}  & \multicolumn{1}{c}{7}      & \multicolumn{1}{c}{28}     & \multicolumn{1}{c}{21} \\
\centering-9.5 & -8   & \multicolumn{1}{c}{0}  & \multicolumn{1}{c}{0}      & \multicolumn{1}{c}{4}      & \multicolumn{1}{c}{4} \\
\bottomrule
\end{tabular}
\end{sc}
\end{small}
\end{center}
\vskip -0.1in
\end{table*}

\subsection{Molecular diversity of the candidate pan-inhibitors}

To assess the performance of our multi-target generative workflow and the diversity of generated molecules, we begin by examining the evolution of three key metrics across each Chemical AL cycle: validity, uniqueness, and novelty. Validity (the percentage of chemically valid molecules among those generated) shows considerable variation across cycles, generally fluctuating between 40\% and 70\% (on average 59.62\% $\pm$ 11.96). This variability can be explained by the stochastic nature of the generation. In contrast, uniqueness (the proportion of non-duplicate valid molecules) and novelty (the percentage of unique molecules not found in the cumulative specific set) remain consistently high throughout the workflow, on average 98.51\% $\pm$ 3.34 and 99.48\% $\pm$ 0.60, respectively (\cref{figure4}, \cref{appendixE}). These results indicate that our multi-target generative model maintains high levels of molecular novelty and uniqueness over successive cycles.

To better understand the efficiency of the AL workflow, we analysed the percentage of molecules passing the two AL level filters across each Affinity AL cycle (\cref{figure5}, \cref{appendixE}). On average, 51.54\% $\pm$ 4.24 of generated molecules pass the Chemical AL filters. This indicates consistent performance throughout the generation process, excluding the first Affinity AL cycle, where only 31.6\% passed due to more restrictive Chemical AL filters (a lower Tanimoto threshold of 0.4 versus 0.6 in subsequent cycles). In contrast, the percentage of molecules passing the Affinity AL filters is less consistent, with an average of 1.33\% $\pm$ 1.64. Notably, there is a downward trend in molecules passing the Affinity filters, likely driven by the linear decay applied to the docking score thresholds.

To further evaluate the structural novelty of the generated molecules, we conducted an additional analysis focusing on the diversity of molecular scaffolds generated across Affinity AL cycles. \cref{figure6}A (\cref{appendixF}) shows that our initial attempt to maximise variability by running a long and stringent Affinity AL cycle (with 40 Chemical AL cycles instead of 20 and a Tanimoto threshold of $<$ 0.4 instead of $<$ 0.6) was successfully achieved, with scaffold diversity increasing from 153 to 654 scaffold clusters. Although the plot appears to plateau around the sixth Affinity AL cycle, this does not imply a complete stop in scaffold generation. In fact, when we consider only the molecules that satisfied the final affinity thresholds of -8 kcal/mol (global and individual) (\cref{figure6}B, \cref{appendixF}), we observe that the number of scaffold clusters continued to grow over time. This indicates that the model was still exploring new and diverse chemical regions within the high-affinity range.

Additionally, the expansion of chemical space throughout the generation was assessed using a UMAP representation of the generated molecules at each Affinity AL cycle (\cref{figure7}, \cref{appendixF}). In the first large Affinity AL cycle, designed to promote variability, the model explored a broad and dispersed region of chemical space. Subsequent, shorter Affinity AL cycles showed more localised exploration, concentrating around previously visited regions, aligning with the design intent of the generation strategy. Notably, the candidate pan-inhibitors are evenly distributed across all chemical space rather than clustered in a single region, indicating that they originate from different and variable scaffolds.

Finally, the candidate pan-inhibitors were searched across the ultra-large chemical libraries Enamine REAL DB ($\approx$6.7B molecules) \cite{noauthor_real_nodate}, ZINC22 ($\approx$37B molecules) \cite{tingle_zinc-22free_2023}, and WuXi ($\approx$3.4B molecules) \cite{noauthor_library-chemistry_nodate}. Among all generated candidates (969), only one molecule had an exact match in WuXi (generated in the regular configuration).

\section{Discussion}

We developed a multi-target generative workflow based on a VAE, iteratively refined through a two-level AL cycle. The first level, the Chemical AL cycle, promotes the generation of diverse, drug-like, and synthetically accessible molecules. The second, the Affinity AL cycle, guides generation toward molecules with multi-target affinity. To evaluate the workflow,  we aimed to generate a diverse set of molecules with high simultaneous predicted affinity toward three homologous targets to obtain potential pan-inhibitor molecules. These targets were the main proteases (Mpro) of SARS-CoV-2, SARS-CoV, and MERS-CoV. This case study, involving structurally similar targets, provided an ideal test to validate the approach's multi-target applicability.

To assess the trade-off between restrictive filtering and flexibility in chemical space exploration, we compared two generative configurations: regular and ablated. For the Mpro targets, the ablated configuration, excluding SMARTS filtering during the Chemical AL cycle and instead applying them post-generation, proved more effective. This configuration enabled broader exploration by not discarding molecules prematurely, leading to a higher yield of pan-inhibitor candidates. However, in other target contexts, the ablated configuration may lead to the accumulation of undesirable motifs that, while chemically unfavourable, could enhance affinity (a property prioritised by the workflow). In such cases, this issue will be exacerbated by the iterative nature of the multi-target generative process, which will propagate such motifs across generations. In situations where post-generation filtering significantly reduces candidates, it may be preferable to use the regular configuration with SMARTS filtering integrated into the chemical AL cycle.

Our diversity analysis demonstrated that the multi-target generative workflow successfully achieved its goal of enhancing molecular variability through the exploration of unseen regions of the chemical space. Broad exploration was prioritised during an initial, extended Affinity AL cycle with stricter similarity thresholds, enabling the generation of a diverse molecular pool. This pool was then refined for multi-target affinity in subsequent, shorter Affinity AL cycles. In these, similarity constraints were relaxed, and affinity thresholds became increasingly stringent. Notably, the candidate pan-inhibitors were evenly distributed across chemical space, highlighting the model’s ability to generate structurally diverse molecules with strong multi-target affinity. Remarkably, nearly all of these (959 out of 960) were absent from several ultra-large chemical libraries, which collectively contain 37 billion molecules (without accounting for duplicates).

The application of our multi-target generative workflow on the discovery of a pan-inhibitor for coronaviruses led to the selection of 960 pan-inhibitor candidates. Nevertheless, while these results are encouraging, further \emph{in silico} and experimental validation will be essential to narrow down our selection and identify candidates whose biological efficacy and safety can be confirmed. These molecules could serve as a starting point for addressing future coronavirus outbreaks.

These findings highlight the potential of our generative workflow in multi-target drug discovery. It is particularly well-suited for polypharmacological strategies aimed at treating complex diseases, where the simultaneous modulation of multiple targets could be transformative. In future applications, the workflow could also be extended to include not only multiple therapeutic targets but also antitargets, enabling simultaneous optimisation of efficacy and safety by actively avoiding undesirable interactions. Besides, accounting for potential antitarget interactions early in the drug discovery process will reduce the risk of costly downstream failures. However, as the number of targets and antitargets increases, so does the computational burden, particularly in the Affinity AL cycle, where docking calculations scale with each additional target. Therefore, computational efficiency through parallelization will be a critical consideration for the broader deployment of this approach.

\section*{Acknowledgements}
This work has been supported by the predoctoral fellowship AGAUR-FI (2024 FI-1 00920) Joan Oró, funded by the Department of Research and Universities of the Government of Catalonia, with co-funding from the European Social Fund Plus. Additional funding was provided by the project CPP2022-009737, financed by MICIU/AEI /10.13039/501100011033 and by the European Union NextGenerationEU/PRTR.

\section*{Impact Statement}

The impact of this work stems from the controllability and flexibility of the presented multi-target generative workflow. Control is achieved by sampling from the continuous latent space of the Seq2Seq VAE, followed by property refinement through a two-level AL workflow designed in close collaboration with medicinal chemistry experts. This allows the generation of chemically viable molecules with desired properties. Model flexibility, enables its application across a broad range of use cases. These include single-target inhibitor design, polypharmacology, pan-inhibitor development, and  designing inhibitors for multiple conformations of a single target. Furthermore, structural variability can be reversed to favor the generation of molecules similar to the specific dataset, making the workflow well-suited for lead optimization. Additionally, the workflow can be used in low-data regimes using a virtual seed. This can be molecules obtained from virtual screening of ultra-large chemical libraries, which can guide generation when known inhibitors are unavailable. Ultimately, these capabilities contribute to accelerating the early-stage of the drug discovery process, and consequently, have a societal impact by reducing the cost and time of the overall drug development process.


\bibliography{references}

\begin{thebibliography}{56}
\providecommand{\natexlab}[1]{#1}
\providecommand{\url}[1]{\texttt{#1}}
\expandafter\ifx\csname urlstyle\endcsname\relax
  \providecommand{\doi}[1]{doi: #1}\else
  \providecommand{\doi}{doi: \begingroup \urlstyle{rm}\Url}\fi

\bibitem[Ahmad et~al.(2024)Ahmad, Bano, Sharma, Sakina, Ahmad, and
  Raza]{ahmad_generative_2024}
Ahmad, S., Bano, N., Sharma, S., Sakina, S., Ahmad, N., and Raza, K.
\newblock Generative {AI} in {Drug} {Designing}: {Current} {State}-of-the-{Art}
  and {Perspectives}.
\newblock In Raza, K., Ahmad, N., and Singh, D. (eds.), \emph{Generative {AI}:
  {Current} {Trends} and {Applications}}, pp.\  427--463. Springer Nature,
  Singapore, 2024.
\newblock ISBN 978-981-97-8460-8.
\newblock \doi{10.1007/978-981-97-8460-8_20}.
\newblock URL \url{https://doi.org/10.1007/978-981-97-8460-8_20}.

\bibitem[Baell \& Holloway(2010)Baell and Holloway]{baell_new_2010}
Baell, J.~B. and Holloway, G.~A.
\newblock New {Substructure} {Filters} for {Removal} of {Pan} {Assay}
  {Interference} {Compounds} ({PAINS}) from {Screening} {Libraries} and for
  {Their} {Exclusion} in {Bioassays}.
\newblock \emph{Journal of Medicinal Chemistry}, 53\penalty0 (7):\penalty0
  2719--2740, April 2010.
\newblock ISSN 0022-2623.
\newblock \doi{10.1021/jm901137j}.
\newblock URL \url{https://doi.org/10.1021/jm901137j}.
\newblock Publisher: American Chemical Society.

\bibitem[Bajusz et~al.(2015)Bajusz, R\'acz, and H\'eberger]{bajusz_why_2015}
Bajusz, D., R\'acz, A., and H\'eberger, K.
\newblock Why is {Tanimoto} index an appropriate choice for fingerprint-based
  similarity calculations?
\newblock \emph{Journal of Cheminformatics}, 7\penalty0 (1):\penalty0 20, May
  2015.
\newblock ISSN 1758-2946.
\newblock \doi{10.1186/s13321-015-0069-3}.
\newblock URL \url{https://doi.org/10.1186/s13321-015-0069-3}.

\bibitem[Bento et~al.(2014)Bento, Gaulton, Hersey, Bellis, Chambers, Davies,
  Kr\"uger, Light, Mak, McGlinchey, Nowotka, Papadatos, Santos, and
  Overington]{bento_chembl_2014}
Bento, A.~P., Gaulton, A., Hersey, A., Bellis, L.~J., Chambers, J., Davies, M.,
  Kr\"uger, F.~A., Light, Y., Mak, L., McGlinchey, S., Nowotka, M., Papadatos,
  G., Santos, R., and Overington, J.~P.
\newblock The {ChEMBL} bioactivity database: an update.
\newblock \emph{Nucleic Acids Research}, 42\penalty0 (Database issue):\penalty0
  D1083--1090, January 2014.
\newblock ISSN 1362-4962.
\newblock \doi{10.1093/nar/gkt1031}.

\bibitem[Bickerton et~al.(2012)Bickerton, Paolini, Besnard, Muresan, and
  Hopkins]{bickerton_quantifying_2012}
Bickerton, G.~R., Paolini, G.~V., Besnard, J., Muresan, S., and Hopkins, A.~L.
\newblock Quantifying the chemical beauty of drugs.
\newblock \emph{Nature Chemistry}, 4\penalty0 (2):\penalty0 90--98, February
  2012.
\newblock ISSN 1755-4349.
\newblock \doi{10.1038/nchem.1243}.
\newblock URL \url{https://www.nature.com/articles/nchem.1243}.
\newblock Publisher: Nature Publishing Group.

\bibitem[Blaschke et~al.(2020)Blaschke, Engkvist, Bajorath, and
  Chen]{blaschke_memory-assisted_2020}
Blaschke, T., Engkvist, O., Bajorath, J., and Chen, H.
\newblock Memory-assisted reinforcement learning for diverse molecular de novo
  design.
\newblock \emph{Journal of Cheminformatics}, 12\penalty0 (1):\penalty0 68,
  November 2020.
\newblock ISSN 1758-2946.
\newblock \doi{10.1186/s13321-020-00473-0}.
\newblock URL \url{https://doi.org/10.1186/s13321-020-00473-0}.

\bibitem[Brenk et~al.(2008)Brenk, Schipani, James, Krasowski, Gilbert,
  Frearson, and Wyatt]{brenk_lessons_2008}
Brenk, R., Schipani, A., James, D., Krasowski, A., Gilbert, I.~H., Frearson,
  J., and Wyatt, P.~G.
\newblock Lessons {Learnt} from {Assembling} {Screening} {Libraries} for {Drug}
  {Discovery} for {Neglected} {Diseases}.
\newblock \emph{ChemMedChem}, 3\penalty0 (3):\penalty0 435--444, 2008.
\newblock ISSN 1860-7187.
\newblock \doi{10.1002/cmdc.200700139}.
\newblock URL
  \url{https://onlinelibrary.wiley.com/doi/abs/10.1002/cmdc.200700139}.

\bibitem[Cicho\'nska et~al.(2024)Cicho\'nska, Ravikumar, and
  Rahman]{cichonska_ai_2024}
Cicho\'nska, A., Ravikumar, B., and Rahman, R.
\newblock {AI} for targeted polypharmacology: {The} next frontier in drug
  discovery.
\newblock \emph{Current Opinion in Structural Biology}, 84:\penalty0 102771,
  February 2024.
\newblock ISSN 0959-440X.
\newblock \doi{10.1016/j.sbi.2023.102771}.
\newblock URL
  \url{https://www.sciencedirect.com/science/article/pii/S0959440X23002452}.

\bibitem[Doveston et~al.(2014)Doveston, Tosatti, Dow, Foley, Li, Campbell,
  House, Churcher, Marsden, and Nelson]{doveston_unified_2014}
Doveston, R.~G., Tosatti, P., Dow, M., Foley, D.~J., Li, H.~Y., Campbell,
  A.~J., House, D., Churcher, I., Marsden, S.~P., and Nelson, A.
\newblock A unified lead-oriented synthesis of over fifty molecular scaffolds.
\newblock \emph{Organic \& Biomolecular Chemistry}, 13\penalty0 (3):\penalty0
  859--865, December 2014.
\newblock ISSN 1477-0539.
\newblock \doi{10.1039/C4OB02287D}.
\newblock URL
  \url{https://pubs.rsc.org/en/content/articlelanding/2015/ob/c4ob02287d}.
\newblock Publisher: The Royal Society of Chemistry.

\bibitem[Edgar(2004)]{edgar_muscle_2004}
Edgar, R.~C.
\newblock {MUSCLE}: multiple sequence alignment with high accuracy and high
  throughput.
\newblock \emph{Nucleic Acids Research}, 32\penalty0 (5):\penalty0 1792--1797,
  2004.
\newblock ISSN 0305-1048.
\newblock \doi{10.1093/nar/gkh340}.
\newblock URL \url{https://www.ncbi.nlm.nih.gov/pmc/articles/PMC390337/}.

\bibitem[Enamine(2024)]{noauthor_real_nodate}
Enamine.
\newblock {REAL} {Database} - {Enamine}, 2024.
\newblock URL
  \url{https://enamine.net/compound-collections/real-compounds/real-database}.

\bibitem[Ertl \& Schuffenhauer(2009)Ertl and
  Schuffenhauer]{ertl_estimation_2009}
Ertl, P. and Schuffenhauer, A.
\newblock Estimation of synthetic accessibility score of drug-like molecules
  based on molecular complexity and fragment contributions.
\newblock \emph{Journal of Cheminformatics}, 1\penalty0 (1):\penalty0 8, June
  2009.
\newblock ISSN 1758-2946.
\newblock \doi{10.1186/1758-2946-1-8}.
\newblock URL \url{https://doi.org/10.1186/1758-2946-1-8}.

\bibitem[Ester et~al.(1996)Ester, Kriegel, and Xu]{ester_density-based_nodate}
Ester, M., Kriegel, H.-P., and Xu, X.
\newblock A {Density}-{Based} {Algorithm} for {Discovering} {Clusters} in
  {Large} {Spatial} {Databases} with {Noise}.
\newblock \emph{KDD-96 Proceedings}, 1996.

\bibitem[Filella-Merce et~al.(2023)Filella-Merce, Molina, Orzechowski, D\'iaz,
  Zhu, Mor, Malo, Yekkirala, Ray, and Guallar]{filella-merce_optimizing_2023}
Filella-Merce, I., Molina, A., Orzechowski, M., D\'iaz, L., Zhu, Y.~M., Mor,
  J.~V., Malo, L., Yekkirala, A.~S., Ray, S., and Guallar, V.
\newblock Optimizing {Drug} {Design} by {Merging} {Generative} {AI} {With}
  {Active} {Learning} {Frameworks}, May 2023.
\newblock URL \url{http://arxiv.org/abs/2305.06334}.
\newblock arXiv:2305.06334 [q-bio].

\bibitem[Gangwal et~al.(2024)Gangwal, Ansari, Ahmad, Azad, and
  Wan~Sulaiman]{gangwal_current_2024}
Gangwal, A., Ansari, A., Ahmad, I., Azad, A.~K., and Wan~Sulaiman, W. M.~A.
\newblock Current strategies to address data scarcity in artificial
  intelligence-based drug discovery: {A} comprehensive review.
\newblock \emph{Computers in Biology and Medicine}, 179:\penalty0 108734,
  September 2024.
\newblock ISSN 0010-4825.
\newblock \doi{10.1016/j.compbiomed.2024.108734}.
\newblock URL
  \url{https://www.sciencedirect.com/science/article/pii/S0010482524008199}.

\bibitem[Grisoni et~al.(2021)Grisoni, Huisman, Button, Moret, Atz, Merk, and
  Schneider]{grisoni_combining_2021}
Grisoni, F., Huisman, B. J.~H., Button, A.~L., Moret, M., Atz, K., Merk, D.,
  and Schneider, G.
\newblock Combining generative artificial intelligence and on-chip synthesis
  for de novo drug design.
\newblock \emph{Science Advances}, 7\penalty0 (24):\penalty0 eabg3338, June
  2021.
\newblock \doi{10.1126/sciadv.abg3338}.
\newblock URL \url{https://www.science.org/doi/10.1126/sciadv.abg3338}.
\newblock Publisher: American Association for the Advancement of Science.

\bibitem[Gupta et~al.(2018)Gupta, M\"uller, Huisman, Fuchs, Schneider, and
  Schneider]{gupta_generative_2018}
Gupta, A., M\"uller, A.~T., Huisman, B. J.~H., Fuchs, J.~A., Schneider, P., and
  Schneider, G.
\newblock Generative {Recurrent} {Networks} for {De} {Novo} {Drug} {Design}.
\newblock \emph{Molecular Informatics}, 37\penalty0 (1-2):\penalty0 1700111,
  January 2018.
\newblock ISSN 1868-1743.
\newblock \doi{10.1002/minf.201700111}.
\newblock URL \url{https://www.ncbi.nlm.nih.gov/pmc/articles/PMC5836943/}.

\bibitem[G\'omez-Bombarelli et~al.(2018)G\'omez-Bombarelli, Wei, Duvenaud,
  Hern\'andez-Lobato, S\'anchez-Lengeling, Sheberla, Aguilera-Iparraguirre,
  Hirzel, Adams, and Aspuru-Guzik]{gomez-bombarelli_automatic_2018}
G\'omez-Bombarelli, R., Wei, J.~N., Duvenaud, D., Hern\'andez-Lobato, J.~M.,
  S\'anchez-Lengeling, B., Sheberla, D., Aguilera-Iparraguirre, J., Hirzel,
  T.~D., Adams, R.~P., and Aspuru-Guzik, A.
\newblock Automatic {Chemical} {Design} {Using} a {Data}-{Driven} {Continuous}
  {Representation} of {Molecules}.
\newblock \emph{ACS Central Science}, 4\penalty0 (2):\penalty0 268--276,
  February 2018.
\newblock ISSN 2374-7943.
\newblock \doi{10.1021/acscentsci.7b00572}.
\newblock URL \url{https://doi.org/10.1021/acscentsci.7b00572}.
\newblock Publisher: American Chemical Society.

\bibitem[Haddad et~al.(2025)Haddad, Litsa, Liu, Yu, Burkhardt, and
  Bhisetti]{haddad_targeted_2025}
Haddad, R., Litsa, E.~E., Liu, Z., Yu, X., Burkhardt, D., and Bhisetti, G.
\newblock Targeted molecular generation with latent reinforcement learning.
\newblock \emph{Scientific Reports}, 15\penalty0 (1):\penalty0 15202, April
  2025.
\newblock ISSN 2045-2322.
\newblock \doi{10.1038/s41598-025-99785-0}.
\newblock URL \url{https://www.nature.com/articles/s41598-025-99785-0}.
\newblock Publisher: Nature Publishing Group.

\bibitem[Halgren(2009)]{halgren_identifying_2009}
Halgren, T.~A.
\newblock Identifying and {Characterizing} {Binding} {Sites} and {Assessing}
  {Druggability}.
\newblock \emph{Journal of Chemical Information and Modeling}, 49\penalty0
  (2):\penalty0 377--389, February 2009.
\newblock ISSN 1549-9596.
\newblock \doi{10.1021/ci800324m}.
\newblock URL \url{https://doi.org/10.1021/ci800324m}.
\newblock Publisher: American Chemical Society.

\bibitem[Halgren et~al.(2004)Halgren, Murphy, Friesner, Beard, Frye, Pollard,
  and Banks]{halgren_glide_2004}
Halgren, T.~A., Murphy, R.~B., Friesner, R.~A., Beard, H.~S., Frye, L.~L.,
  Pollard, W.~T., and Banks, J.~L.
\newblock Glide:  {A} {New} {Approach} for {Rapid}, {Accurate} {Docking} and
  {Scoring}. 2. {Enrichment} {Factors} in {Database} {Screening}.
\newblock \emph{Journal of Medicinal Chemistry}, 47\penalty0 (7):\penalty0
  1750--1759, March 2004.
\newblock ISSN 0022-2623.
\newblock \doi{10.1021/jm030644s}.
\newblock URL \url{https://doi.org/10.1021/jm030644s}.
\newblock Publisher: American Chemical Society.

\bibitem[Huang et~al.(2023)Huang, Shuai, Qiao, Hou, Zeng, Xia, Xie, Fang, Li,
  Yoon, Huang, Hu, You, Quan, Zhao, Guo, Zhang, Ma, Zhang, Wang, Yang, Zhang,
  Nan, Xu, Wang, Lei, Chu, and Yang]{huang_new_2023}
Huang, C., Shuai, H., Qiao, J., Hou, Y., Zeng, R., Xia, A., Xie, L., Fang, Z.,
  Li, Y., Yoon, C., Huang, Q., Hu, B., You, J., Quan, B., Zhao, X., Guo, N.,
  Zhang, S., Ma, R., Zhang, J., Wang, Y., Yang, R., Zhang, S., Nan, J., Xu, H.,
  Wang, F., Lei, J., Chu, H., and Yang, S.
\newblock A new generation {Mpro} inhibitor with potent activity against
  {SARS}-{CoV}-2 {Omicron} variants.
\newblock \emph{Signal Transduction and Targeted Therapy}, 8\penalty0
  (1):\penalty0 1--13, March 2023.
\newblock ISSN 2059-3635.
\newblock \doi{10.1038/s41392-023-01392-w}.
\newblock URL \url{https://www.nature.com/articles/s41392-023-01392-w}.
\newblock Publisher: Nature Publishing Group.

\bibitem[Isigkeit et~al.(2024)Isigkeit, H\"ormann, Schallmayer, Scholz, Lillich,
  Ehrler, Hufnagel, B\"uchner, Marschner, Pabel, Proschak, and
  Merk]{isigkeit_automated_2024}
Isigkeit, L., H\"ormann, T., Schallmayer, E., Scholz, K., Lillich, F.~F.,
  Ehrler, J. H.~M., Hufnagel, B., B\"uchner, J., Marschner, J.~A., Pabel, J.,
  Proschak, E., and Merk, D.
\newblock Automated design of multi-target ligands by generative deep learning.
\newblock \emph{Nature Communications}, 15\penalty0 (1):\penalty0 7946,
  September 2024.
\newblock ISSN 2041-1723.
\newblock \doi{10.1038/s41467-024-52060-8}.
\newblock URL \url{https://www.nature.com/articles/s41467-024-52060-8}.
\newblock Publisher: Nature Publishing Group.

\bibitem[Jadhav et~al.(2010)Jadhav, Ferreira, Klumpp, Mott, Austin, Inglese,
  Thomas, Maloney, Shoichet, and Simeonov]{jadhav_quantitative_2010}
Jadhav, A., Ferreira, R.~S., Klumpp, C., Mott, B.~T., Austin, C.~P., Inglese,
  J., Thomas, C.~J., Maloney, D.~J., Shoichet, B.~K., and Simeonov, A.
\newblock Quantitative {Analyses} of {Aggregation}, {Autofluorescence}, and
  {Reactivity} {Artifacts} in a {Screen} for {Inhibitors} of a {Thiol}
  {Protease}.
\newblock \emph{Journal of Medicinal Chemistry}, 53\penalty0 (1):\penalty0
  37--51, January 2010.
\newblock ISSN 0022-2623.
\newblock \doi{10.1021/jm901070c}.
\newblock URL \url{https://doi.org/10.1021/jm901070c}.
\newblock Publisher: American Chemical Society.

\bibitem[Katoh \& Standley(2013)Katoh and Standley]{katoh_mafft_2013}
Katoh, K. and Standley, D.~M.
\newblock {MAFFT} {Multiple} {Sequence} {Alignment} {Software} {Version} 7:
  {Improvements} in {Performance} and {Usability}.
\newblock \emph{Molecular Biology and Evolution}, 30\penalty0 (4):\penalty0
  772--780, April 2013.
\newblock ISSN 0737-4038.
\newblock \doi{10.1093/molbev/mst010}.
\newblock URL \url{https://www.ncbi.nlm.nih.gov/pmc/articles/PMC3603318/}.

\bibitem[Kim et~al.(2025)Kim, Chen, Cheng, Gindulyte, He, He, Li, Shoemaker,
  Thiessen, Yu, Zaslavsky, Zhang, and Bolton]{kim_pubchem_2025}
Kim, S., Chen, J., Cheng, T., Gindulyte, A., He, J., He, S., Li, Q., Shoemaker,
  B., Thiessen, P., Yu, B., Zaslavsky, L., Zhang, J., and Bolton, E.
\newblock {PubChem} 2025 update.
\newblock \emph{Nucleic Acids Research}, 53\penalty0 (D1):\penalty0
  D1516--D1525, January 2025.
\newblock ISSN 1362-4962.
\newblock \doi{10.1093/nar/gkae1059}.
\newblock URL \url{https://doi.org/10.1093/nar/gkae1059}.

\bibitem[Knox et~al.(2024)Knox, Wilson, Klinger, Franklin, Oler, Wilson, Pon,
  Cox, Chin, Strawbridge, Garcia-Patino, Kruger, Sivakumaran, Sanford, Doshi,
  Khetarpal, Fatokun, Doucet, Zubkowski, Rayat, Jackson, Harford, Anjum, Zakir,
  Wang, Tian, Lee, Liigand, Peters, Wang, Nguyen, So, Sharp, da Silva,
  Gabriel, Scantlebury, Jasinski, Ackerman, Jewison, Sajed, Gautam, and
  Wishart]{knox_drugbank_2024}
Knox, C., Wilson, M., Klinger, C., Franklin, M., Oler, E., Wilson, A., Pon, A.,
  Cox, J., Chin, N.~E., Strawbridge, S., Garcia-Patino, M., Kruger, R.,
  Sivakumaran, A., Sanford, S., Doshi, R., Khetarpal, N., Fatokun, O., Doucet,
  D., Zubkowski, A., Rayat, D., Jackson, H., Harford, K., Anjum, A., Zakir, M.,
  Wang, F., Tian, S., Lee, B., Liigand, J., Peters, H., Wang, R.~Q., Nguyen,
  T., So, D., Sharp, M., da Silva, R., Gabriel, C., Scantlebury, J., Jasinski,
  M., Ackerman, D., Jewison, T., Sajed, T., Gautam, V., and Wishart, D.
\newblock {DrugBank} 6.0: the {DrugBank} {Knowledgebase} for 2024.
\newblock \emph{Nucleic Acids Research}, 52\penalty0 (D1):\penalty0
  D1265--D1275, January 2024.
\newblock ISSN 0305-1048.
\newblock \doi{10.1093/nar/gkad976}.
\newblock URL \url{https://doi.org/10.1093/nar/gkad976}.

\bibitem[Korshunova et~al.(2022)Korshunova, Huang, Capuzzi, Radchenko, Savych,
  Moroz, Wells, Willson, Tropsha, and Isayev]{korshunova_generative_2022}
Korshunova, M., Huang, N., Capuzzi, S., Radchenko, D.~S., Savych, O., Moroz,
  Y.~S., Wells, C.~I., Willson, T.~M., Tropsha, A., and Isayev, O.
\newblock Generative and reinforcement learning approaches for the automated de
  novo design of bioactive compounds.
\newblock \emph{Communications Chemistry}, 5\penalty0 (1):\penalty0 1--11,
  October 2022.
\newblock ISSN 2399-3669.
\newblock \doi{10.1038/s42004-022-00733-0}.
\newblock URL \url{https://www.nature.com/articles/s42004-022-00733-0}.
\newblock Publisher: Nature Publishing Group.

\bibitem[Kyro et~al.(2024)Kyro, Morgunov, Brent, and
  Batista]{kyro_chemspaceal_2024}
Kyro, G.~W., Morgunov, A., Brent, R.~I., and Batista, V.~S.
\newblock {ChemSpaceAL}: {An} {Efficient} {Active} {Learning} {Methodology}
  {Applied} to {Protein}-{Specific} {Molecular} {Generation}.
\newblock \emph{Journal of Chemical Information and Modeling}, 64\penalty0
  (3):\penalty0 653--665, February 2024.
\newblock ISSN 1549-9596.
\newblock \doi{10.1021/acs.jcim.3c01456}.
\newblock URL \url{https://doi.org/10.1021/acs.jcim.3c01456}.
\newblock Publisher: American Chemical Society.

\bibitem[Landrum et~al.(2025)Landrum, Tosco, Kelley, Rodriguez, Cosgrove,
  Vianello, sriniker, Gedeck, Jones, NadineSchneider, Kawashima, Nealschneider,
  Dalke, Swain, Cole, Turk, Savelev, tadhurst cdd, Vaucher, W\'ojcikowski, Take,
  Walker, Scalfani, Faara, Ujihara, Probst, Lehtivarjo, godin, Pahl, and
  Monat]{landrum_rdkitrdkit_2025}
Landrum, G., Tosco, P., Kelley, B., Rodriguez, R., Cosgrove, D., Vianello, R.,
  sriniker, Gedeck, P., Jones, G., NadineSchneider, Kawashima, E.,
  Nealschneider, D., Dalke, A., Swain, M., Cole, B., Turk, S., Savelev, A.,
  tadhurst cdd, Vaucher, A., W\'ojcikowski, M., Take, I., Walker, R., Scalfani,
  V.~F., Faara, H., Ujihara, K., Probst, D., Lehtivarjo, J., godin, g., Pahl,
  A., and Monat, J.
\newblock rdkit/rdkit: 2025\_03\_1 ({Q1} 2025) {Release}, March 2025.
\newblock URL \url{https://zenodo.org/records/15115844}.

\bibitem[Liu et~al.(2021)Liu, Ye, van Vlijmen, Emmerich, IJzerman, and van
  Westen]{liu_drugex_2021}
Liu, X., Ye, K., van Vlijmen, H. W.~T., Emmerich, M. T.~M., IJzerman, A.~P.,
  and van Westen, G. J.~P.
\newblock {DrugEx} v2: de novo design of drug molecules by {Pareto}-based
  multi-objective reinforcement learning in polypharmacology.
\newblock \emph{Journal of Cheminformatics}, 13\penalty0 (1):\penalty0 85,
  November 2021.
\newblock ISSN 1758-2946.
\newblock \doi{10.1186/s13321-021-00561-9}.
\newblock URL \url{https://doi.org/10.1186/s13321-021-00561-9}.

\bibitem[Loeffler et~al.(2024)Loeffler, Wan, Kl\"ahn, Bhati, and
  Coveney]{loeffler_optimal_2024}
Loeffler, H.~H., Wan, S., Kl\"ahn, M., Bhati, A.~P., and Coveney, P.~V.
\newblock Optimal {Molecular} {Design}: {Generative} {Active} {Learning}
  {Combining} {REINVENT} with {Precise} {Binding} {Free} {Energy} {Ranking}
  {Simulations}.
\newblock \emph{Journal of Chemical Theory and Computation}, 20\penalty0
  (18):\penalty0 8308--8328, September 2024.
\newblock ISSN 1549-9618.
\newblock \doi{10.1021/acs.jctc.4c00576}.
\newblock URL \url{https://doi.org/10.1021/acs.jctc.4c00576}.
\newblock Publisher: American Chemical Society.

\bibitem[Madhavi~Sastry et~al.(2013)Madhavi~Sastry, Adzhigirey, Day,
  Annabhimoju, and Sherman]{madhavi_sastry_protein_2013}
Madhavi~Sastry, G., Adzhigirey, M., Day, T., Annabhimoju, R., and Sherman, W.
\newblock Protein and ligand preparation: parameters, protocols, and influence
  on virtual screening enrichments.
\newblock \emph{Journal of Computer-Aided Molecular Design}, 27\penalty0
  (3):\penalty0 221--234, March 2013.
\newblock ISSN 1573-4951.
\newblock \doi{10.1007/s10822-013-9644-8}.
\newblock URL \url{https://doi.org/10.1007/s10822-013-9644-8}.

\bibitem[McInnes et~al.(2020)McInnes, Healy, and Melville]{mcinnes_umap_2020-1}
McInnes, L., Healy, J., and Melville, J.
\newblock {UMAP}: {Uniform} {Manifold} {Approximation} and {Projection} for
  {Dimension} {Reduction}, September 2020.
\newblock URL \url{http://arxiv.org/abs/1802.03426}.
\newblock arXiv:1802.03426 [stat].

\bibitem[MedChemExpress(2024)]{noauthor_medchemexpress_nodate}
MedChemExpress.
\newblock {MedChemExpress}: {Master} of {Bioactive} {Molecules} {\textbar}
  {Inhibitors}, {Screening} {Libraries} \& {Proteins}, 2024.
\newblock URL \url{https://www.medchemexpress.com/}.

\bibitem[Munson et~al.(2024)Munson, Chen, Bogosian, Kreisberg, Licon, Abagyan,
  Kuenzi, and Ideker]{munson_novo_2024}
Munson, B.~P., Chen, M., Bogosian, A., Kreisberg, J.~F., Licon, K., Abagyan,
  R., Kuenzi, B.~M., and Ideker, T.
\newblock De novo generation of multi-target compounds using deep generative
  chemistry.
\newblock \emph{Nature Communications}, 15\penalty0 (1):\penalty0 3636, May
  2024.
\newblock ISSN 2041-1723.
\newblock \doi{10.1038/s41467-024-47120-y}.
\newblock URL \url{https://www.nature.com/articles/s41467-024-47120-y}.
\newblock Publisher: Nature Publishing Group.

\bibitem[Neumann \& Klein(2025)Neumann and Klein]{neumann_benchmark_2025}
Neumann, A. and Klein, R.
\newblock A {Benchmark} {Set} of {Bioactive} {Molecules} for {Diversity}
  {Analysis} of {Compound} {Libraries} and {Combinatorial} {Chemical} {Spaces},
  April 2025.
\newblock URL
  \url{https://chemrxiv.org/engage/chemrxiv/article-details/67ebaa8c81d2151a028aed21}.

\bibitem[Odaibo(2019)]{odaibo_tutorial_2019}
Odaibo, S.
\newblock Tutorial: {Deriving} the {Standard} {Variational} {Autoencoder}
  ({VAE}) {Loss} {Function}, July 2019.
\newblock URL \url{http://arxiv.org/abs/1907.08956}.
\newblock arXiv:1907.08956 [cs].

\bibitem[Olivecrona et~al.(2017)Olivecrona, Blaschke, Engkvist, and
  Chen]{olivecrona_molecular_2017}
Olivecrona, M., Blaschke, T., Engkvist, O., and Chen, H.
\newblock Molecular de-novo design through deep reinforcement learning.
\newblock \emph{Journal of Cheminformatics}, 9\penalty0 (1):\penalty0 48,
  September 2017.
\newblock ISSN 1758-2946.
\newblock \doi{10.1186/s13321-017-0235-x}.
\newblock URL \url{https://doi.org/10.1186/s13321-017-0235-x}.

\bibitem[rdkit.Chem package(2025)]{noauthor_rdkitchem_nodate}
rdkit.Chem package.
\newblock rdkit.{Chem} package --- {The} {RDKit} 2025.03.1 documentation, 2025.
\newblock URL \url{https://www.rdkit.org/docs/source/rdkit.Chem.html}.

\bibitem[Schr\"odinger(2025)]{noauthor_ligprep_nodate}
Schr\"odinger, L.
\newblock {Schr\"odinger Release 2025-2: LigPrep}, 2025.
\newblock URL \url{https://www.schrodinger.com/platform/products/ligprep/}.

\bibitem[scikit learn(2025)]{noauthor_dbscan_nodate}
scikit learn.
\newblock {DBSCAN}, 2025.
\newblock URL
  \url{https://scikit-learn/stable/modules/generated/sklearn.cluster.DBSCAN.html}.

\bibitem[Shahhamzehei et~al.(2022)Shahhamzehei, Abdelfatah, and
  Efferth]{shahhamzehei_silico_2022}
Shahhamzehei, N., Abdelfatah, S., and Efferth, T.
\newblock In {Silico} and {In} {Vitro} {Identification} of {Pan}-{Coronaviral}
  {Main} {Protease} {Inhibitors} from a {Large} {Natural} {Product} {Library}.
\newblock \emph{Pharmaceuticals}, 15\penalty0 (3):\penalty0 308, March 2022.
\newblock ISSN 1424-8247.
\newblock \doi{10.3390/ph15030308}.
\newblock URL \url{https://www.mdpi.com/1424-8247/15/3/308}.
\newblock Number: 3 Publisher: Multidisciplinary Digital Publishing Institute.

\bibitem[Sheikholeslami et~al.(2025)Sheikholeslami, Mazrouei, Gheisari, Fasihi,
  Irajpour, and Motahharynia]{sheikholeslami_druggen_2025}
Sheikholeslami, M., Mazrouei, N., Gheisari, Y., Fasihi, A., Irajpour, M., and
  Motahharynia, A.
\newblock {DrugGen} enhances drug discovery with large language models and
  reinforcement learning.
\newblock \emph{Scientific Reports}, 15\penalty0 (1):\penalty0 13445, April
  2025.
\newblock ISSN 2045-2322.
\newblock \doi{10.1038/s41598-025-98629-1}.
\newblock URL \url{https://www.nature.com/articles/s41598-025-98629-1}.
\newblock Publisher: Nature Publishing Group.

\bibitem[Sievers \& Higgins(2018)Sievers and Higgins]{sievers_clustal_2018}
Sievers, F. and Higgins, D.~G.
\newblock Clustal {Omega} for making accurate alignments of many protein
  sequences.
\newblock \emph{Protein Science}, 27\penalty0 (1):\penalty0 135--145, 2018.
\newblock ISSN 1469-896X.
\newblock \doi{10.1002/pro.3290}.
\newblock URL \url{https://onlinelibrary.wiley.com/doi/abs/10.1002/pro.3290}.
\newblock \_eprint: https://onlinelibrary.wiley.com/doi/pdf/10.1002/pro.3290.

\bibitem[Swanson et~al.(2024)Swanson, Liu, Catacutan, Arnold, Zou, and
  Stokes]{swanson_generative_2024}
Swanson, K., Liu, G., Catacutan, D.~B., Arnold, A., Zou, J., and Stokes, J.~M.
\newblock Generative {AI} for designing and validating easily synthesizable and
  structurally novel antibiotics.
\newblock \emph{Nature Machine Intelligence}, 6\penalty0 (3):\penalty0
  338--353, March 2024.
\newblock ISSN 2522-5839.
\newblock \doi{10.1038/s42256-024-00809-7}.
\newblock URL \url{https://www.nature.com/articles/s42256-024-00809-7}.

\bibitem[Tang et~al.(2024)Tang, Dai, Knight, Wu, Li, Li, and
  Gerstein]{tang_survey_2024}
Tang, X., Dai, H., Knight, E., Wu, F., Li, Y., Li, T., and Gerstein, M.
\newblock A {Survey} of {Generative} {AI} for de novo {Drug} {Design}: {New}
  {Frontiers} in {Molecule} and {Protein} {Generation}, June 2024.
\newblock URL \url{http://arxiv.org/abs/2402.08703}.
\newblock arXiv:2402.08703 [q-bio].

\bibitem[Tingle et~al.(2023)Tingle, Tang, Castanon, Gutierrez, Khurelbaatar,
  Dandarchuluun, Moroz, and Irwin]{tingle_zinc-22free_2023}
Tingle, B.~I., Tang, K.~G., Castanon, M., Gutierrez, J.~J., Khurelbaatar, M.,
  Dandarchuluun, C., Moroz, Y.~S., and Irwin, J.~J.
\newblock {ZINC}-22-{A} {Free} {Multi}-{Billion}-{Scale} {Database} of
  {Tangible} {Compounds} for {Ligand} {Discovery}.
\newblock \emph{Journal of Chemical Information and Modeling}, 63\penalty0
  (4):\penalty0 1166--1176, February 2023.
\newblock ISSN 1549-9596.
\newblock \doi{10.1021/acs.jcim.2c01253}.
\newblock URL \url{https://doi.org/10.1021/acs.jcim.2c01253}.
\newblock Publisher: American Chemical Society.

\bibitem[tmtools(2025)]{noauthor_tmtools_nodate}
tmtools.
\newblock tmtools: {Python} bindings around the {TM}-align code for structural
  alignment of proteins, 2025.
\newblock URL \url{https://github.com/jvkersch/tmtools}.

\bibitem[umap learn(2025)]{noauthor_umap_nodate}
umap learn.
\newblock {UMAP}: {Uniform} {Manifold} {Approximation} and {Projection} for
  {Dimension} {Reduction} --- umap 0.5.8 documentation, 2025.
\newblock URL \url{https://umap-learn.readthedocs.io/en/latest/}.

\bibitem[Van~Tilborg \& Grisoni(2024)Van~Tilborg and
  Grisoni]{van_tilborg_traversing_2024}
Van~Tilborg, D. and Grisoni, F.
\newblock Traversing {Chemical} {Space} with {Active} {Deep} {Learning}: {A}
  {Computational} {Framework} for {Low}-data {Drug} {Discovery}, February 2024.
\newblock URL
  \url{https://chemrxiv.org/engage/chemrxiv/article-details/65d8833ce9ebbb4db9098cb5}.

\bibitem[Walters(2025)]{walters_patwaltersrd_filters_2025}
Walters, P.
\newblock {PatWalters}/rd\_filters, March 2025.
\newblock URL \url{https://github.com/PatWalters/rd_filters}.
\newblock original-date: 2018-08-08T01:29:21Z.

\bibitem[Waskom et~al.(2017)Waskom, Botvinnik, O'Kane, Hobson, Lukauskas,
  Gemperline, Augspurger, Halchenko, Cole, Warmenhoven, Ruiter, Pye, Hoyer,
  Vanderplas, Villalba, Kunter, Quintero, Bachant, Martin, Meyer, Miles, Ram,
  Yarkoni, Williams, Evans, Fitzgerald, Brian, Fonnesbeck, Lee, and
  Qalieh]{waskom_mwaskomseaborn_2017}
Waskom, M., Botvinnik, O., O'Kane, D., Hobson, P., Lukauskas, S., Gemperline,
  D.~C., Augspurger, T., Halchenko, Y., Cole, J.~B., Warmenhoven, J., Ruiter,
  J.~d., Pye, C., Hoyer, S., Vanderplas, J., Villalba, S., Kunter, G.,
  Quintero, E., Bachant, P., Martin, M., Meyer, K., Miles, A., Ram, Y.,
  Yarkoni, T., Williams, M.~L., Evans, C., Fitzgerald, C., Brian, Fonnesbeck,
  C., Lee, A., and Qalieh, A.
\newblock mwaskom/seaborn: v0.8.1 ({September} 2017), September 2017.
\newblock URL \url{https://zenodo.org/records/883859}.

\bibitem[WuXi(2024)]{noauthor_library-chemistry_nodate}
WuXi.
\newblock Library-{Chemistry}, 2024.
\newblock URL \url{https://chemistry.wuxiapptec.com/library}.

\bibitem[{wwPDB consortium}(2019)]{wwpdb_consortium_protein_2019}
{wwPDB consortium}.
\newblock Protein {Data} {Bank}: the single global archive for {3D}
  macromolecular structure data.
\newblock \emph{Nucleic Acids Research}, 47\penalty0 (D1):\penalty0 D520--D528,
  January 2019.
\newblock ISSN 0305-1048.
\newblock \doi{10.1093/nar/gky949}.
\newblock URL \url{https://doi.org/10.1093/nar/gky949}.

\bibitem[Zhavoronkov et~al.(2019)Zhavoronkov, Ivanenkov, Aliper, Veselov,
  Aladinskiy, Aladinskaya, Terentiev, Polykovskiy, Kuznetsov, Asadulaev,
  Volkov, Zholus, Shayakhmetov, Zhebrak, Minaeva, Zagribelnyy, Lee, Soll,
  Madge, Xing, Guo, and Aspuru-Guzik]{zhavoronkov_deep_2019}
Zhavoronkov, A., Ivanenkov, Y.~A., Aliper, A., Veselov, M.~S., Aladinskiy,
  V.~A., Aladinskaya, A.~V., Terentiev, V.~A., Polykovskiy, D.~A., Kuznetsov,
  M.~D., Asadulaev, A., Volkov, Y., Zholus, A., Shayakhmetov, R.~R., Zhebrak,
  A., Minaeva, L.~I., Zagribelnyy, B.~A., Lee, L.~H., Soll, R., Madge, D.,
  Xing, L., Guo, T., and Aspuru-Guzik, A.
\newblock Deep learning enables rapid identification of potent {DDR1} kinase
  inhibitors.
\newblock \emph{Nature Biotechnology}, 37\penalty0 (9):\penalty0 1038--1040,
  September 2019.
\newblock ISSN 1546-1696.
\newblock \doi{10.1038/s41587-019-0224-x}.
\newblock URL \url{https://www.nature.com/articles/s41587-019-0224-x}.
\newblock Publisher: Nature Publishing Group.

\end{thebibliography}
\bibliographystyle{icml2025}

\newpage
\appendix
\onecolumn

\section{Datasets}
\label{appendixA}

The general training set was obtained from the ChEMBL 30 database \cite{bento_chembl_2014} (2.7 million molecules) to which a drug-likeness filter was applied to ensure suitability for the generative model (molecular weight between 150 and 500 Da, free of salts, and compliant with Lipinski's Rule of Five), reducing its size to 247,199 molecules in SMILES format.

The fixed specific training set was assembled by collecting experimentally validated inhibitors (with known $IC_{50}$ affinity values) targeting the Mpro of SARS-CoV-2, SARS-CoV, and MERS-CoV. These molecules were retrieved from multiple chemical databases, including PDB \cite{wwpdb_consortium_protein_2019}, ChEMBL \cite{bento_chembl_2014}, PubChem \cite{kim_pubchem_2025}, DrugBank \cite{knox_drugbank_2024}, and MedChemExpress \cite{noauthor_medchemexpress_nodate}, resulting in a total of 477 molecules after removing peptides. Of this, 267 corresponded to SARS-CoV-2 inhibitors, 198 to SARS-CoV inhibitors, and 30 to MERS-CoV inhibitors. Given this imbalance between species, all retrieved molecules were subjected to a cross-docking protocol against the three targets to create a balanced specific training set. Molecules below a predicted affinity threshold of -5.9 kcal/mol (docking score from Glide) to all three proteases were retained. This docking score threshold was established by calculating the average docking score of all molecules across all targets. This process yielded a final specific training set comprising 214 unique molecules, each with a minimum signal of predicted multi-target affinity. 

\section{Target Selection and Preparation}
\label{appendixB}

All available crystallographic structures of the main protease (Mpro) from the Protein Data Bank (PDB) \cite{wwpdb_consortium_protein_2019} were collected and classified according to their viral origin: 479 from SARS-CoV-2, 32 from SARS-CoV, and 32 from MERS-CoV. Target structures were preprocessed by removing water molecules, ligands, and ions. In cases where structures contained multiple chains, reflecting both Mpro dimeric and monomeric states, individual chains were separated, and only monomeric chains within a defined residue range (minimum of 290 and maximum of 330) encompassing the catalytic site were retained, as the objective was to inhibit a single active monomer. For each virus, all structures were superimposed onto a designated reference structure using TMalign \cite{noauthor_tmtools_nodate}, and further prepared using Schrödinger Protein Preparation Wizard \cite{madhavi_sastry_protein_2013}, which completed the missing atoms and optimized the structural geometry. 

To remove redundant structures (structures with identical catalytic site conformation), we cluster them based on the 3D volume of their catalytic site. To do so, we computed the 3D catalytic site volumes using Schrödinger SiteMap \cite{halgren_identifying_2009} and constructed a pairwise catalytic site volume overlapping matrix. Then we hierarchically clustered this matrix using Seaborn \cite{waskom_mwaskomseaborn_2017}. This allowed us to identify clusters of very similar 3D catalytic site volumes from which we extract representatives, reducing the total number of Mpro structures from 543 to 195.

Cross-docking of the fixed specific set was subsequently conducted with these 195 structures, and receptor selection was based on the total number of inhibitors (the structure with the highest counts was selected) that were below the docking score threshold of -5.9 kcal/mol (same used in the construction of the specific set). Further evaluation based on resolution, model completeness, and electron density quality was done in the selected structures. Based on these criteria, the final selected target structures were 7RNW for SARS-CoV-2, 2GX4 for SARS-CoV, and 7ENE for MERS-CoV.

To assess the degree of similarity among homologous Mpro proteins across different coronavirus species, a sequence conservation analysis was performed. Pairwise sequence alignments were carried out using three independent tools, MAFFT \cite{katoh_mafft_2013}, MUSCLE \cite{edgar_muscle_2004}, and Clustal Omega \cite{sievers_clustal_2018}, all of which produced consistent results. The resulting pairwise sequence percentage identities (\%ID) are summarised in \cref{table2}. In addition to sequence-based comparisons, we superimposed the target structures using TM-align \cite{noauthor_tmtools_nodate}. Pairwise root-mean-square deviations (RMSD) reflecting high structural conservation are also reported in \cref{table2}. From the previous structural superimpositions, we extracted their structure-based sequence alignments, from which we derive the \%ID of the catalytic sites,  the key regions targeted in the design of multi-target inhibitors (\cref{table3}).

\begin{table}[H]
\caption{Percentage identity and RMSD of the Mpro sequences for SARS-CoV-2 (SARS2), SARS-CoV (SARS), and MERS-CoV (MERS).}
\label{table2}
\vskip 0.15in
\begin{center}
\begin{small}
\begin{sc}
\begin{tabular}{lccc}
\toprule
\%ID and RMSD & SARS2 & SARS & MERS \\
\midrule
SARS2  &            -       & 96.08\% (RMSD=1.06) & 50\% (RMSD=1.32)    \\
SARS   & 96.08\% (RMSD=1.06)&              -      & 50.98\% (RMSD=1.55) \\
MERS   & 50\% (RMSD=1.32)   & 50.98\% (RMSD=1.55) &         -           \\

\bottomrule
\end{tabular}
\end{sc}
\end{small}
\end{center}
\vskip -0.1in
\end{table}

\begin{table}[H]
\caption{Percentage identity derived from the structure-based sequence alignment just on the catalytic site of the Mpro for SARS-CoV-2 (SARS2), SARS-CoV (SARS), and MERS-CoV (MERS).}
\label{table3}
\vskip 0.15in
\begin{center}
\begin{small}
\begin{sc}
\begin{tabular}{lcccr}
\toprule
\%ID & SARS2 & SARS & MERS \\
\midrule
SARS2  &     -   & 96.7\%  & 63.33\% \\
SARS   & 96.7\%  &     -   & 66.67\% \\
MERS   & 63.33\% & 66.67\% &     -   \\

\bottomrule
\end{tabular}
\end{sc}
\end{small}
\end{center}
\vskip -0.1in
\end{table}

\section{Docking Protocol}
\label{appendixC}

Docking simulations were computed using Schrödinger’s Glide software \cite{halgren_glide_2004}, operating in standard precision (SP) mode, with a grid centred on the catalytic active site (C145:SG) and fixed dimensions of 10\r{A} for the inner box and 30\r{A} for the outer box. For each molecule, up to five docking poses were evaluated, and no structural constraints were applied.

\section{Cumulative Counts across Affinity AL Cycles}
\label{appendixD}

\begin{figure}[H]
\vskip 0.2in
\begin{center}
\centerline{\includegraphics[width=0.5\columnwidth]{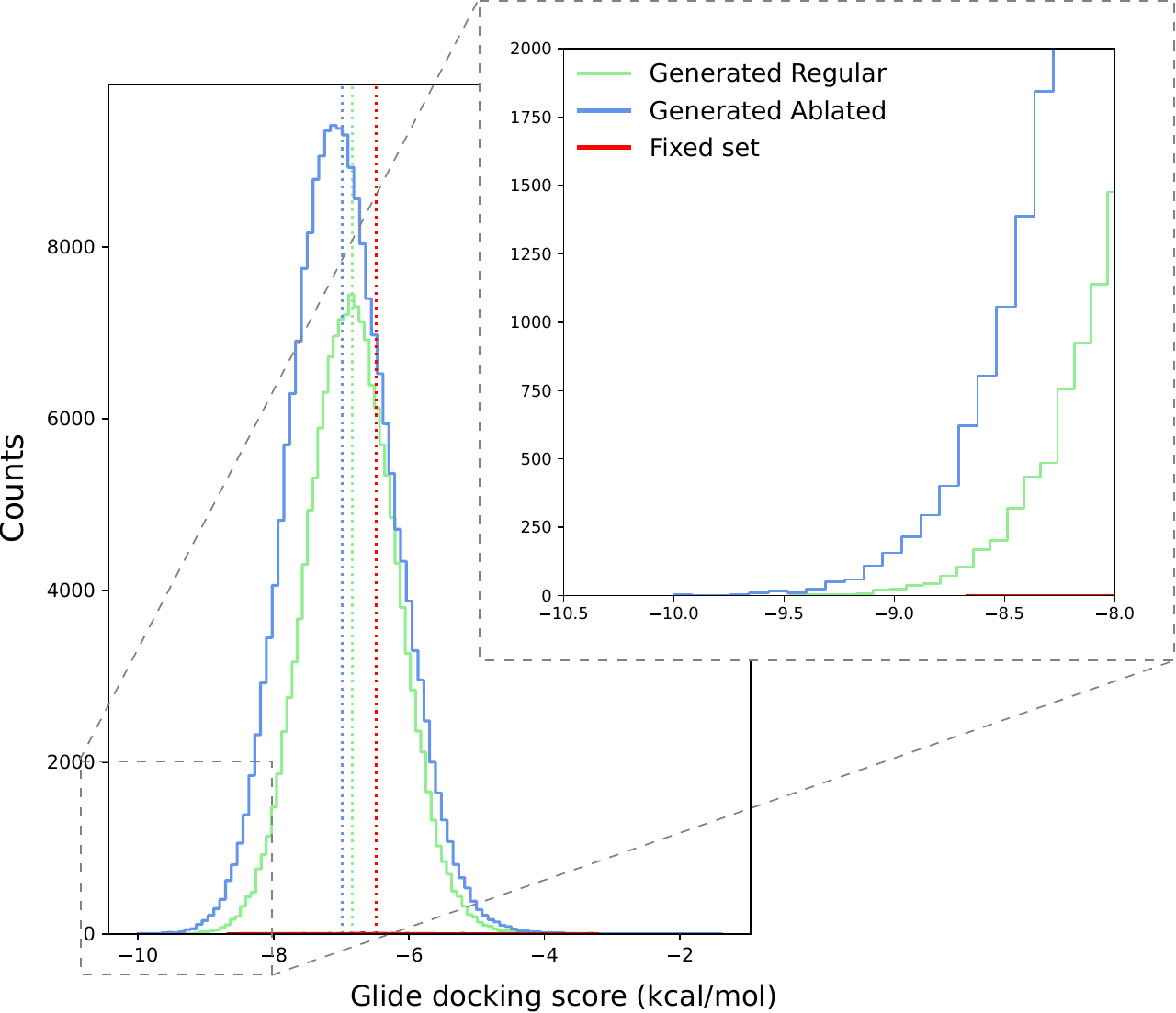}}
\caption{Global docking score cumulative histograms of generated molecules across all Affinity AL cycles for the regular and the ablated configurations, and also the fixed specific set. The zoomed-in section represents the molecules under the final defined threshold of -8 kcal/mol.}
\label{figure3}
\end{center}
\vskip -0.2in
\end{figure}

\section{Generation Statistics}
\label{appendixE}

The validity, uniqueness, and novelty of the generated molecules at each Chemical AL cycle were calculated as follows:

\begin{align}
\text{Validity} &= \left( \frac{N_{\text{val}}}{N_{\text{gen}}} \right) \times 100 \tag{7} \\
\text{Uniqueness} &= \left( \frac{N_{\text{uni}}}{N_{\text{val}}} \right) \times 100 \tag{8} \\
\text{Novelty} &= \left( \frac{N_{\text{unk}}}{N_{\text{uni}}} \right) \times 100 \tag{9}
\end{align}

where $N_{gen}$ stands for the total generated molecules, $N_{val}$ for the total valid molecules from $N_{gen}$, $N_{uni}$ for the total unique molecules (non-duplicated) from $N_{val}$, and $N_{unk}$ for the total number of unique molecules not found in the cumulative specific set. 

\begin{figure}[H]
\vskip 0.2in
\begin{center}
\centerline{\includegraphics[width=\columnwidth]{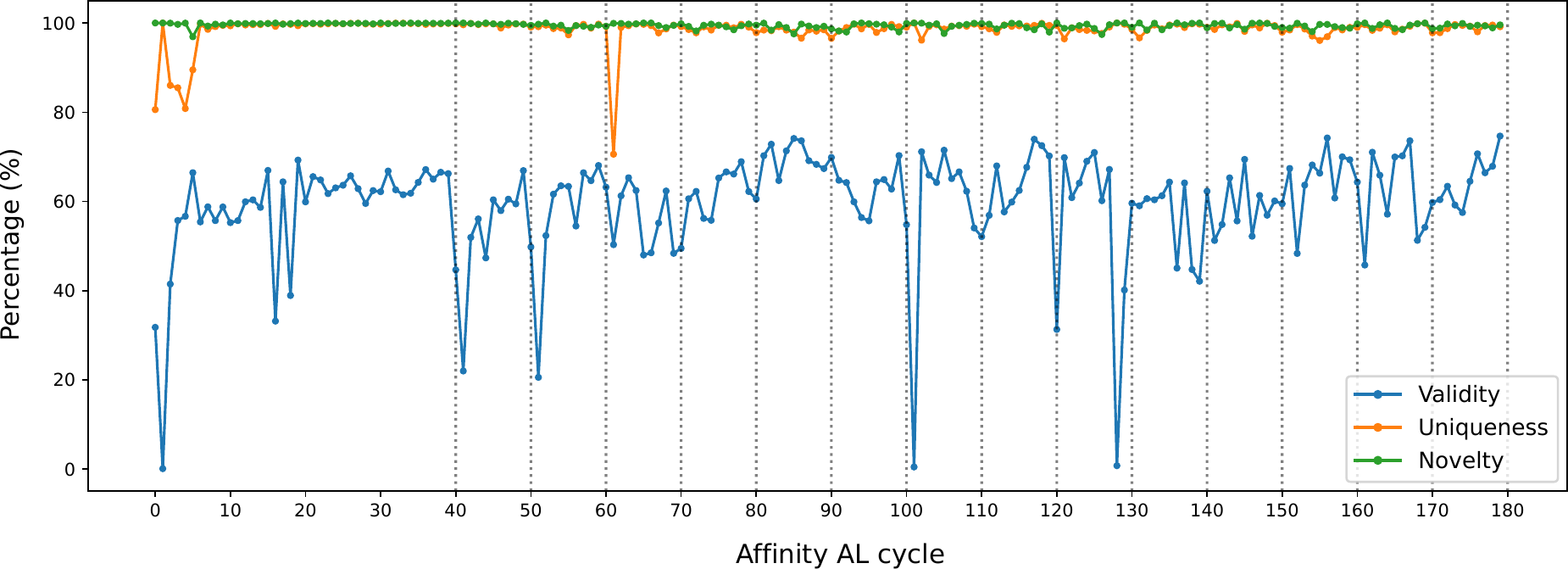}}
\caption{Validity, uniqueness, and novelty of the molecules generated at each Chemical AL cycle. The vertical dotted lines represent the division between Affinity AL cycles.}
\label{figure4}
\end{center}
\vskip -0.2in
\end{figure}

We tracked the number of molecules that fulfilled the criteria of the Chemical and Affinity AL cycles at each Affinity AL iteration. 

\begin{figure}[H]
\vskip 0.2in
\begin{center}
\centerline{\includegraphics[width=0.8\columnwidth]{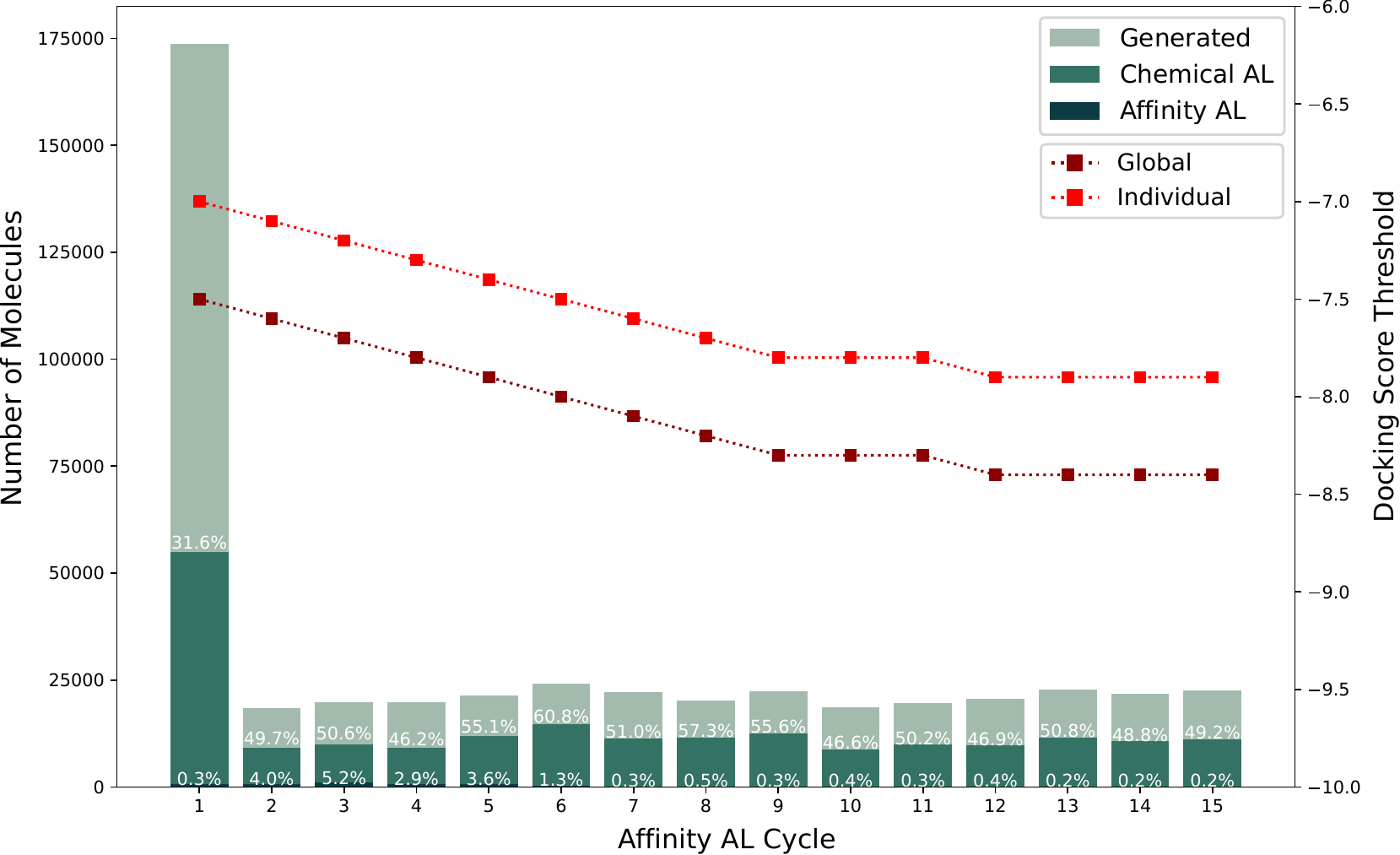}}
\caption{Number and percentage of generated molecules at each Affinity AL cycle fulfilling the Chemical AL filters and the Affinity AL filters. The docking score thresholds (global and individual) that were applied at each Affinity AL cycle are represented on the right-y axis.}
\label{figure5}
\end{center}
\vskip -0.2in
\end{figure}

\section{Molecular Diversity}
\label{appendixF}

To extract scaffolds from molecules that met specified global and individual docking score thresholds, we used the Scaffolds.MurckoScaffold module from Chem package in the RDKit Python library \cite{noauthor_rdkitchem_nodate}. For clustering these molecular scaffolds, we applied the DBSCANS algorithm \cite{ester_density-based_nodate} from the scikit-learn Python library \cite{noauthor_dbscan_nodate}, varying the value of the epsilon parameter (minimum similarity between any pair of scaffolds within the same cluster) to ensure that the observed trends are not artifacts of the chosen epsilon value.

\begin{figure}[H]
\vskip 0.2in
\begin{center}
\centerline{\includegraphics[width=0.9\columnwidth]{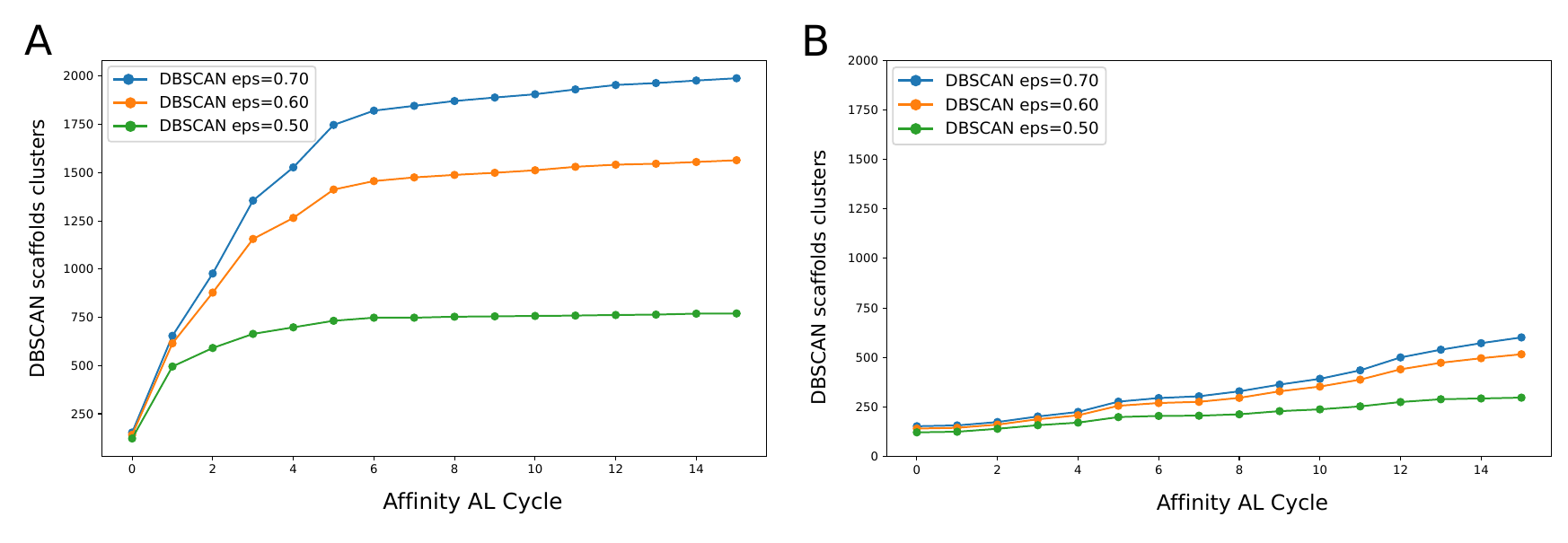}}
\caption{Evolution of scaffold clusters across Affinity AL cycles. Each connected line represents a DBSCAN clustering performed with a different epsilon parameter, as indicated in the legend. A) Evolution of scaffold clusters among generated molecules below the thresholds applied at each Affinity AL cycle, and B) Evolution of scaffold clusters among generated molecules below thresholds of -8 kcal/mol (global and individual).}
\label{figure6}
\end{center}
\vskip -0.2in
\end{figure}

\begin{figure}[H]
\vskip 0.2in
\begin{center}
\centerline{\includegraphics[width=0.7\columnwidth]{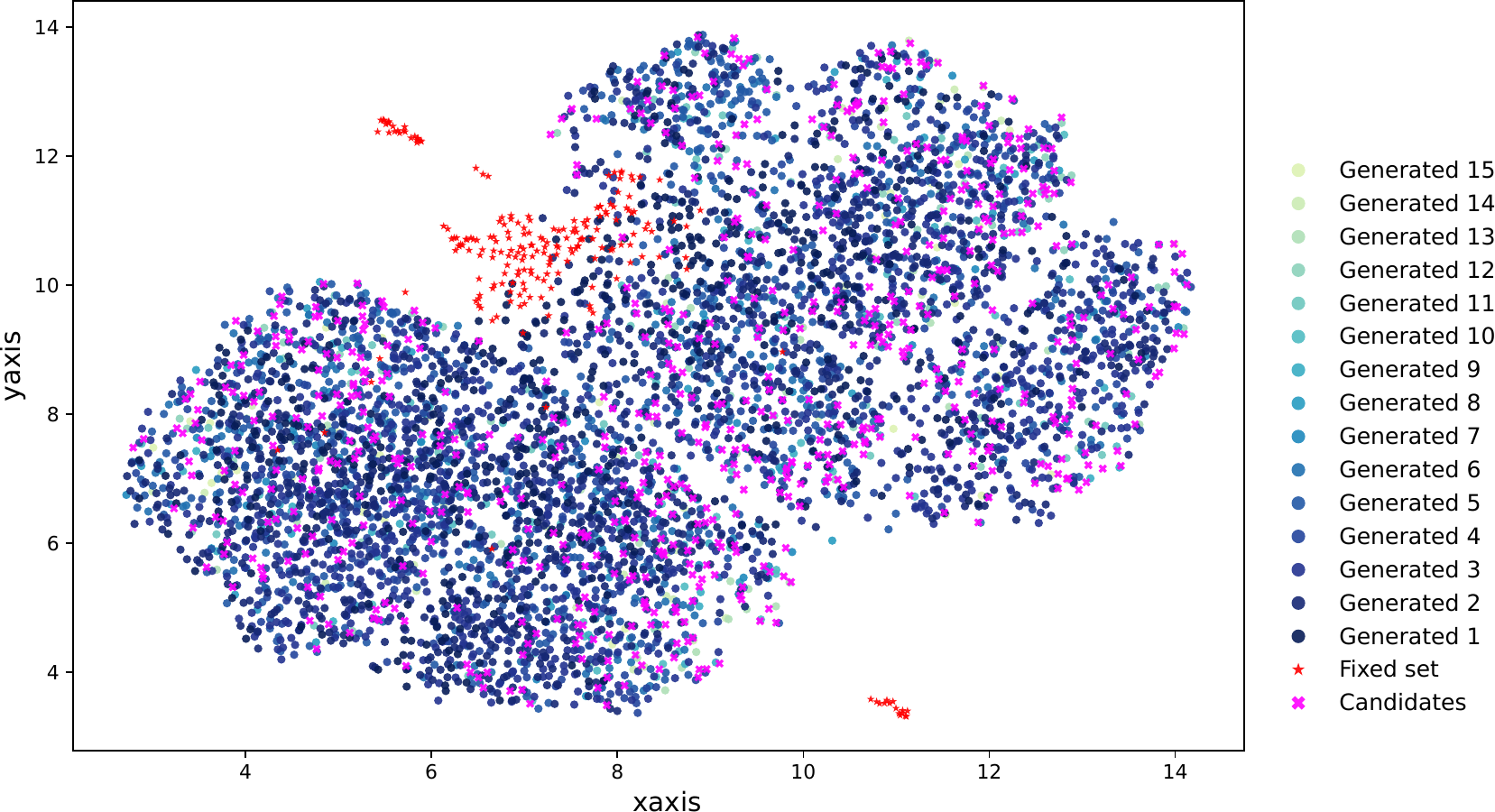}}
\caption{UMAP illustrating the generated molecules over the Affinity AL cycles, with marker styles and colours indicating different cycles as shown in the legend. Fucsin crosses represent the candidate pan-inhibitors, thus generated molecules that passed the docking score thresholds of -8 kcal (global) and -8 kcal/mol (individual).}
\label{figure7}
\end{center}
\vskip -0.2in
\end{figure}

The two-dimensional molecular representation provided by the Uniform Manifold Approximation and Projection (UMAP) algorithm \cite{mcinnes_umap_2020-1} enabled visualisation of the exploratory behaviour of our multi-target generative workflow. The UMAP in \cref{figure7} was obtained with the umap-learn Python library \cite{noauthor_umap_nodate}, using Morgan4 fingerprints and Hamming distance.

\section{Performance time}
\label{appendixG}

To evaluate the computational efficiency of the proposed multi-target generative workflow, execution times were calculated for each major component of the workflow. The presented times are real time (wall-clock time):
\begin{itemize}
\item The training time for the general dataset ($\approx$200k molecules) required 38.26 $\pm$ 0.53 minutes on a single GPU H100.
\item Fine-tuning on a specific dataset of 1,000 molecules, starting from the pretrained weights, required a total of 10.52 $\pm$ 0.73 minutes on a single GPU H1000. It is important to notice that the specific dataset has a cumulative nature, grows with each Chemical AL cycle, which eventually leads to a gradual rise in time over iterations.
\item For generating one molecule in one GPU H100, the model takes 1.51 $\pm$ 0.15 seconds.
\item Chemical AL filters took about 0.05 ± 8.5e-4 seconds per molecule, running on a single core of a 4th Generation Intel Xeon Scalable processor.
\item A Chemical AL cycle, comprising a specific dataset fine-tuning, a generation of 3,500 molecules, and their consecutive chemical AL filtering, required a total of 1 hour and 42 minutes $\pm$ 8 minutes.
\item An affinity AL cycle requires $n$ Chemical AL cycles, followed by LigPrep \cite{noauthor_ligprep_nodate} for ligand preparation and Glide docking \cite{halgren_glide_2004}. The ligprep and docking for 1.000 molecules on 3 targets took 48.58 $\pm$ 1.37 minutes, parallelised on 30 cores. Thus, an Affinity AL cycle with $n=10$ Chemical AL cycles will take approximately 17 hours and 43 minutes.
\item Assuming a regular execution of our multi-target generative workflow, including 10 Affinity AL cycles, the expected time would be of 177 hours and 15 minutes ($\approx$ 7.39 days).
\end{itemize}

\end{document}